\documentclass{arxivpaper}

\usepackage{float}
\usepackage{colortbl}
\usepackage{nicefrac}
\usepackage{multirow}
\usepackage{bm}

\definecolor{gainLow}{HTML}{EDF5F9}
\definecolor{gainMid}{HTML}{D6E9F3}
\definecolor{gainHigh}{HTML}{B8D8E8}
\definecolor{rankStable}{HTML}{EDF5F9}
\definecolor{rankDefault}{HTML}{C5DFEC}
\definecolor{localCertaintyShade}{HTML}{75B3D2}
\newcommand{\localmean}[2]{\cellcolor{localCertaintyShade!#1!white}$#2$}

\newcommand{\rankdefault}[1]{\cellcolor{rankDefault}\textbf{#1}}

\newcommand{\modelicon}[1]{%
  \raisebox{-0.19\height}{\includegraphics[height=1.70ex]{assets/icons/#1}}}
\newcommand{\qwenicon}{\modelicon{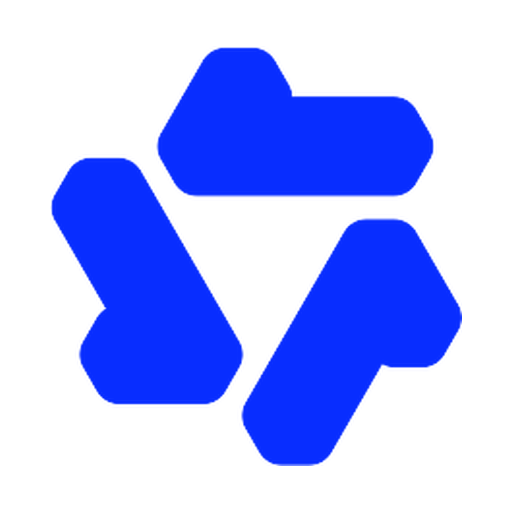}}
\newcommand{\gemmaicon}{\modelicon{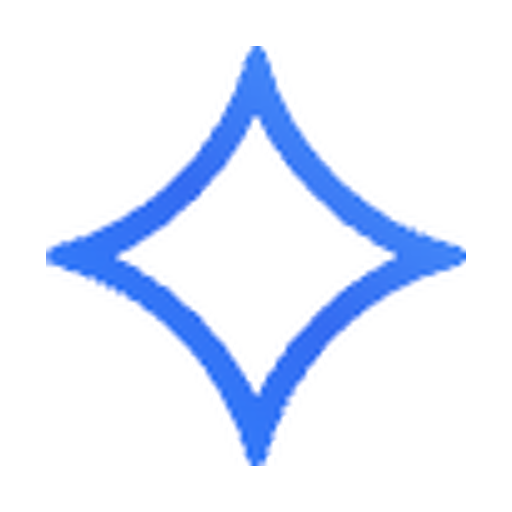}}
\newcommand{\openaiicon}{\modelicon{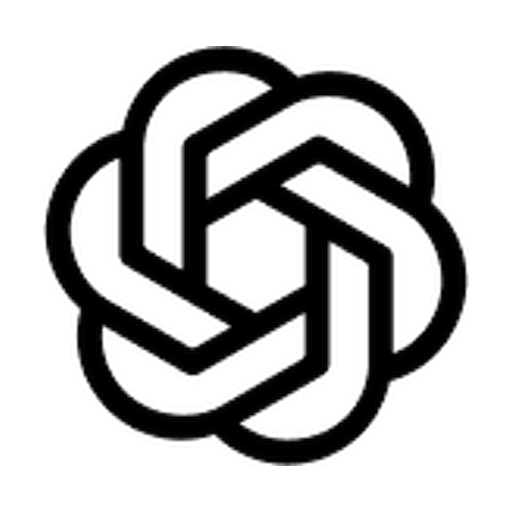}}

\newtcolorbox{keyfinding}[1]{
  enhanced,
  breakable,
  colback=PaperPanel,
  colframe=PaperPrimary,
  borderline west={2.2pt}{0pt}{PaperPrimary},
  boxrule=0.45pt,
  arc=1.8mm,
  left=10pt,right=10pt,top=8pt,bottom=9pt,
  before skip=10pt,after skip=10pt,
  fonttitle=\normalfont\bfseries,
  coltitle=PaperPrimary,
  title={#1},
}

\newtcolorbox{promptbox}[1]{
  enhanced,
  breakable,
  colback=PaperPanel,
  colframe=PaperRule,
  colbacktitle=PaperPrimary,
  coltitle=white,
  title={#1},
  fonttitle=\normalfont\bfseries\small,
  boxrule=0.55pt,
  arc=2mm,
  left=10pt,right=10pt,top=8pt,bottom=9pt,
  titlerule=0pt,
}

\title{Certainty Is Not Just Correctness: Rethinking Token-Level Certainty in LLM Reasoning}
\papershorttitle{Certainty Is Not Just Correctness}
\paperheader{Preprint}
\paperdate{2026-9-3}

\author{Yunfan Zhou\texorpdfstring{\textsuperscript{1}}{}\texorpdfstring{, }{, }
  Ye Zhu\texorpdfstring{\textsuperscript{1}}{}\texorpdfstring{, }{, }
  Zhihai Wang\texorpdfstring{\textsuperscript{2}}{}\texorpdfstring{, }{, }
  Jianguo Yao\texorpdfstring{\textsuperscript{1}}{}\texorpdfstring{, }{, }
  Haibing Guan\texorpdfstring{\textsuperscript{1}}{}\texorpdfstring{, }{, }
  Xijun Li\texorpdfstring{\textsuperscript{1*}}{}}
\affiliation{\textsuperscript{1}School of Computer Science, Shanghai Jiao Tong University\\
  \textsuperscript{2}Qwen Team, Alibaba Group}
\paperemail{\texttt{zhou.yunfan@sjtu.edu.cn, lixijun@sjtu.edu.cn}}
\authornote{* Corresponding author.}

\begin{document}
\flushbottom

\maketitle

\begin{abstract}
Token-level certainty is widely used as a proxy for correctness in LLM training and inference. However, the performance of certainty-based methods depends both on the information in certainty scores and on how those scores are used. We therefore directly assess certainty's predictive ability through controlled empirical evaluations across models and tasks. We distinguish two prediction targets: identifying questions a model is more likely to answer correctly and distinguishing correct from incorrect responses to the same question. In our experiments, certainty is generally better at identifying questions a model is likely to answer correctly than at distinguishing correct from incorrect responses to the same question. Certainty also varies systematically across token types and positions within words, reflecting local properties of words and text form. Information about question difficulty appears early in generation, while the weaker information about answer correctness is more concentrated near the end. These findings show that the information certainty provides for decisions depends on the prediction target, the model, the certainty metric, and which token positions in the response are included in aggregation. We further demonstrate the practical value of these findings for test-time compute. We allocate the number of responses using certainty early in generation and weight answer votes using certainty near the end of each response. Compared with a fixed-sampling majority-voting baseline, this approach increases overall accuracy from 78.71\% to 79.54\% while reducing generated-token cost by 82.4\%.
\end{abstract}

\section{Introduction}
Assessing the correctness of large language model (LLM) reasoning is essential for deciding which outputs to trust and how to improve them. Token-level certainty provides a readily available signal: it summarizes the model's next-token distribution and can be aggregated over a generated response. Using information already produced during generation, certainty enables decisions about model outputs without ground-truth answers or a separately trained verifier \citep{Kang2025Scalable,Fu2025Deep}. This accessibility makes certainty useful for both training and inference.

Across these uses, certainty serves as a proxy for correctness: higher scores are treated as evidence that an output is more likely to be correct. This premise connects response selection, which favors high-scoring candidates \citep{Kang2025Scalable}, to training with certainty rewards, which reinforces high-scoring responses \citep{Zhao2025Learning}. It also motivates selecting actions based on their token probabilities in interactive tasks \citep{Wang2024Soft} and using certainty to decide whether a reasoning path should continue or an answer is ready to return \citep{Fu2025Deep,Nogueira2025Certainty}. Although these decisions use scores differently, they rely on the correctness information expected from certainty.

However, gains from using certainty are measured for the training or inference method as a whole. They therefore do not directly establish how reliably certainty predicts correctness or under which conditions that relationship holds. Previous work has directly evaluated response ranking \citep{Taubenfeld2025Confidence,Wang2024Soft}, compared aggregation rules \citep{Kang2025Scalable,Fu2025Deep}, and controlled reasoning content \citep{Kim2026Reasoning}. A systematic account connecting these measurements is still missing: what certainty predicts, how the information it provides depends on the way certainty is measured, and which uses that information supports. Without such an account, a method's gains offer little guidance on when the same signal can support other decisions. We address this gap through direct, controlled evaluation: we compare certainty scores against correctness labels on the same generated responses across models and tasks.

\begin{figure}[!t]
  \centering
  \includegraphics[trim=0 0.407bp 0 0,clip,width=\linewidth]{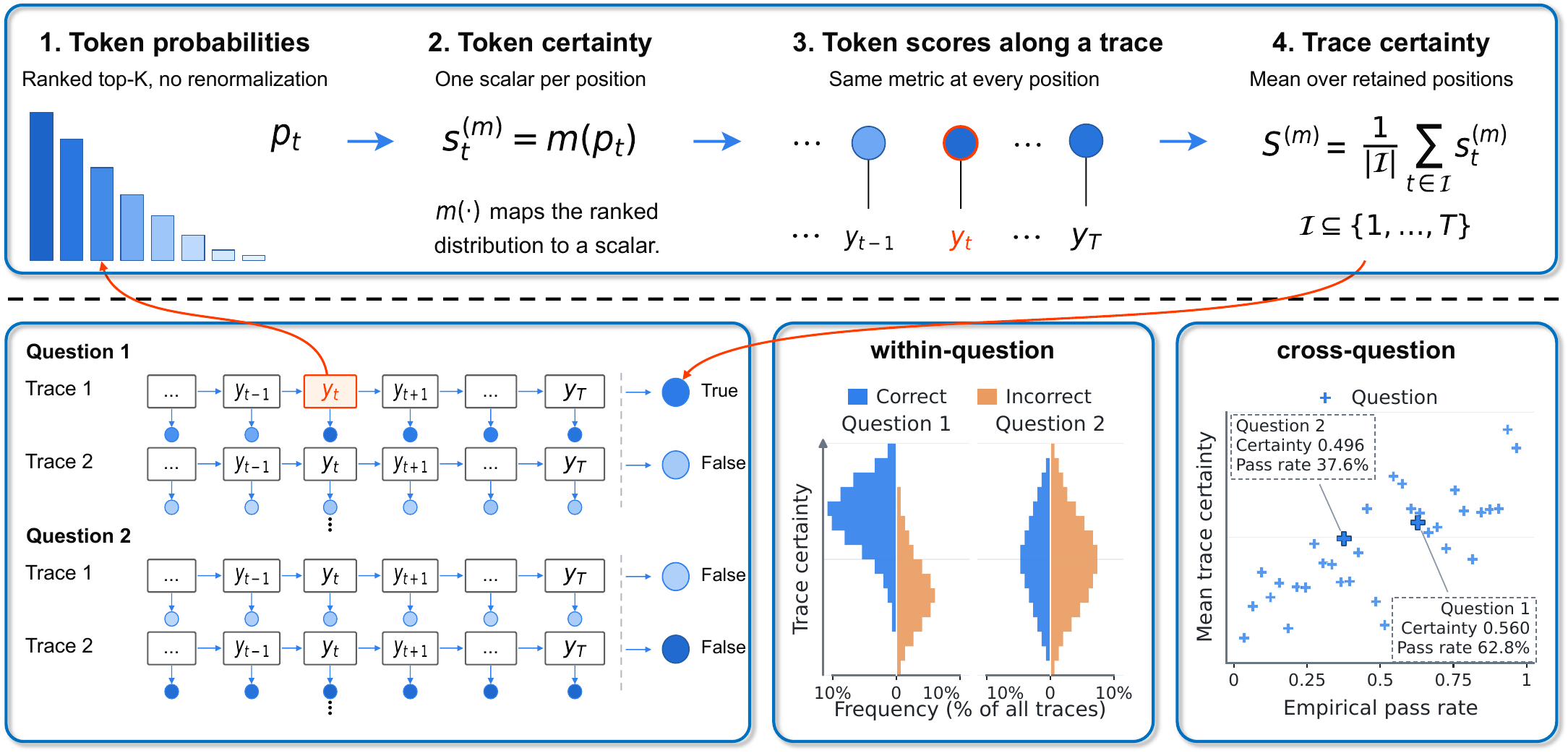}
  \caption{From token-level certainty to two prediction targets. A metric scores each next-token distribution; averaging scores over selected positions yields a response (or trace) score. Within-question comparisons test whether scores distinguish correct from incorrect responses to the same question. Cross-question comparisons test whether a question's mean response score reflects its proportion of correct responses.}
  \label{fig:thesis}
\end{figure}

Our study covers five reasoning models, three mathematical and scientific benchmarks, and four certainty metrics. We distinguish identifying questions a model is more likely to answer correctly from identifying correct responses to the same question (Figure~\ref{fig:thesis}). With token scores averaged over complete responses, all four metrics are better at the former when results are averaged across benchmarks for each model. Information about question difficulty is already available early in generation, whereas the weaker information distinguishing correct and incorrect responses is more concentrated near the end. This difference motivates Target-Aligned Certainty Scaling (TACS). TACS uses scores from the beginning of a response to decide how many responses to generate for a question, and scores near the end to weight their votes for final answer selection. In our experiments, TACS reaches 79.54\% accuracy at 17.6\% of the generated-token cost of fixed 256-response majority voting, whose accuracy is 78.71\%. Early budget allocation reduces token cost; weighting votes using certainty averaged over a final token window improves average accuracy, with gains that depend on the model. These findings show why certainty must be assessed in relation to the information a decision requires and when that information becomes available.

Our main contribution is a systematic empirical account of the information token-level certainty provides for reasoning. We find empirically that, across the models and tasks evaluated, certainty is generally better at identifying questions a model is likely to answer correctly than at distinguishing correct from incorrect responses to the same question. Its predictive ability varies with the model, task, and scoring choices. Certainty also reflects local properties of words and text form, and simply excluding selected tokens does not consistently improve prediction. Information about question difficulty is available early, whereas information about response correctness is more concentrated near the end. These findings suggest choosing which part of a response to score according to the information a decision requires. We examine the practical value of this principle through test-time compute experiments. TACS applies these principles to reduce generation costs, while gains from weighting response votes remain model-dependent.

\section{Related Work}
\label{section:preliminaries}

\paragraph{Token-level certainty.}
Token-level certainty summarizes the concentration of an LLM's next-token
distribution at each generation position.
We use a top-$K$ version of \textbf{self-certainty} that averages the negative
log probabilities of the $K$ most probable tokens
\citep{Kang2025Scalable,Fu2025Deep}.
\textbf{Negative entropy} sums probability-weighted log probabilities
\citep{Kang2025Scalable,Kim2026Reasoning}; \textbf{Maximum probability}
uses the highest next-token probability \citep{Nogueira2025Certainty,Yang2025Dynamic}.
In our comparisons, we use its logarithm, maximum log probability (MaxLP).
For ranked full-softmax probabilities $p_0\geq\cdots\geq p_{K-1}$,
we define these three metrics without renormalizing the top-$K$ slice:
\begin{equation}
\label{eq:certainty-metrics}
\begin{array}{@{}l@{\quad}c@{\quad}l@{}}
\text{Self-certainty }(\mathrm{SC}_K) & = & \displaystyle -\frac{1}{K}\sum_{j=0}^{K-1}\log p_j, \\[6pt]
\text{Top-}K\text{ Negative Entropy }(\mathrm{NegEnt}_K) & = & \displaystyle \sum_{j=0}^{K-1}p_j\log p_j, \\[6pt]
\text{Maximum Log Probability }(\mathrm{MaxLP}) & = & \log p_0.
\end{array}
\end{equation}
\paragraph{From token certainty to trace scores.}
Aggregating token-level scores yields a score for a generated response,
or trace. A typical approach is mean aggregation: token scores $s_t$ are averaged over
a selected set of positions $\mathcal I$:
\begin{equation}
\label{eq:trace-aggregation}
S=\frac{1}{|\mathcal I|}\sum_{t\in\mathcal I}s_t.
\end{equation}
Self-Certainty \citep{Kang2025Scalable} averages scores over the entire response.
DeepConf \citep{Fu2025Deep} also considers the mean over a tail segment and
the sliding window with the lowest mean score. DEER \citep{Yang2025Dynamic}
averages per-position maximum probabilities over an induced trial answer.
RENT \citep{Prabhudesai2025Maximizing} averages token-level negative entropy
over response positions.
These methods extend scoring from complete responses to local segments or trial answers.

\paragraph{From certainty scores to decisions.}
Model-internal probability signals guide certainty-based output selection,
model training, and test-time compute control.
For output selection, scores are used to choose candidate responses or actions
\citep{Kang2025Scalable,Wang2024Soft} and weight votes across responses
\citep{Kang2025Scalable,Taubenfeld2025Confidence}.
For model training, self-certainty \citep{Zhao2025Learning} and negative token
entropy \citep{Prabhudesai2025Maximizing,Agarwal2025Unreasonable} serve as
reinforcement learning rewards. Token entropy also supports selecting update
positions, with policy-gradient updates restricted to high-entropy tokens
in reinforcement learning with verifiable rewards
\citep{wang20258020rulehighentropyminority}.
For test-time compute control, confidence or entropy signals guide reasoning-path
termination and decisions about further reasoning
\citep{Fu2025Deep,Nogueira2025Certainty,Yang2025Dynamic,Sharma2025Think},
as well as adjustments to the number of sampled responses \citep{Kim2026Reliability}.
Using certainty to select, reinforce, or retain responses assumes
that it helps distinguish correct from incorrect outputs.
However, a systematic understanding of
how this discriminative ability varies with the model, task, scoring metric,
and response region remains incomplete.

\section{Evaluation Framework}
\label{sec:evaluation-framework}

The performance of certainty-based methods depends on both certainty scores
and their use; performance alone does not establish how well the scores
distinguish correct from incorrect responses.
For example, consider a typical certainty application: weighted answer
voting, which sums certainty scores for each answer and selects the answer
with the largest total. Suppose four of five responses to a question give
the same incorrect answer and one gives the correct answer. Even if the certainty score is 2 for the correct response and 1 for every
incorrect response, yielding perfect correctness ranking, the incorrect answer still wins
4 to 2. Any permutation of these scores leaves the winner unchanged.
Thus, the same voting outcome can occur with very different ranking quality;
voting accuracy alone cannot evaluate certainty's discriminative ability.
To measure this ability directly, we establish a controlled evaluation
framework that compares scoring choices on the same generated responses and
correctness labels. We separately measure how well scores distinguish correct
from incorrect responses to the same question and how well they distinguish
questions with different empirical accuracies for a fixed model. This distinction lets us identify
what certainty predicts and examine how its discriminative ability varies
with the metric, scoring position, model, and task.

\subsection{Evaluation Objectives}
\label{sec:ranking-and-utility}

When using certainty to distinguish correct from incorrect responses, we must
also ask whether scores distinguish questions or responses to the same question.
Consider two questions with ten responses each. The first has nine correct
responses, and all ten responses score one; the second has one correct response,
and all ten responses score zero. The AUROC is $0.9$ when all twenty responses are combined,
but $0.5$ for each question separately.
The score separates questions but cannot select correct responses within either.
We therefore distinguish discrimination among responses to the same question
from discrimination of relative question difficulty for a fixed model.

For question $q$ with $n_q$ responses, let $S_{qi}$ denote certainty and
$y_{qi}\in\{0,1\}$ correctness. Define
\begin{equation}
\label{eq:question-level-summaries}
\bar S_q=\frac{1}{n_q}\sum_{i=1}^{n_q}S_{qi},
\qquad
\hat r_q=\frac{1}{n_q}\sum_{i=1}^{n_q}y_{qi}.
\end{equation}
Here $\bar S_q$ is mean certainty and $\hat r_q$ is empirical accuracy.
For each question $q$, let $i^+$ and $i^-$ be indices drawn independently and
uniformly from its correct and incorrect responses, respectively.
We measure within-question discrimination between correct and incorrect responses to the same question,
and cross-question discrimination between questions with different empirical accuracies for the same model:
\begin{equation}
\text{within-question AUROC}
=\Pr(S_{qi^+}>S_{qi^-})
+\tfrac12\Pr(S_{qi^+}=S_{qi^-}),
\label{eq:within-question-auroc}
\end{equation}
\begin{equation}
\text{cross-question C-Index}
=\Pr(\bar S_q>\bar S_{q'}\mid\hat r_q>\hat r_{q'})
+\tfrac12\Pr(\bar S_q=\bar S_{q'}\mid\hat r_q>\hat r_{q'}).
\label{eq:cross-question-cindex}
\end{equation}
For within-question AUROC, we average per-question AUROC equally over
questions containing both correct and incorrect responses and exclude
single-class questions. For cross-question C-Index, we weight comparable pairs equally
and exclude pairs tied in empirical accuracy. Both metrics assign half credit to score ties.
Empirical accuracy depends on the model and sampled responses.

\subsection{Experiment Setup}
\label{sec:experiment-setup}

\textbf{Datasets and models.} We evaluate five models---Qwen3-8B/32B \citep{Yang2025Qwen3}, GPT-OSS-20B
\citep{OpenAI2025GptOss}, Gemma 4-E4B, and Gemma 4-26B-A4B \citep{GemmaTeam2026Gemma4}---on
mathematical reasoning with AIME 2024/2025 \citep{Balunovic2025Matharena} and
scientific question answering with GPQA-Diamond \citep{Rein2023Gpqa}.
We refer to Gemma 4-26B-A4B as Gemma4-26B throughout the rest of this paper.
For each model and question, we apply different scoring choices to the same
generated responses and evaluate them against the same correctness labels.
We report results across models and datasets to distinguish shared patterns
from effects that depend on a particular model or task. Detailed comparisons
and settings accompany the corresponding analyses.

We use these evaluation metrics to compare four certainty scores. SC, NegEnt,
and MaxLP from Section~\ref{section:preliminaries} summarize concentration in the
next-token distribution. To examine how certainty varies with candidate rank, we introduce the logit gap.
Let $z_k$ denote the logit corresponding to probability $p_k$, where tokens are
ordered by decreasing probability and indexed from zero. The gap between the
most probable token and the token indexed by $k$ is
\begin{equation}
\label{eq:ranked-logit-gap}
\Delta_k=\log p_0-\log p_k=z_0-z_k,
\quad 1\leq k\leq K-1.
\end{equation}
Varying $k$ allows us to compare the discriminative power of gaps at different
ranks. Our main four-metric comparison uses the top 20 token probabilities,
with logit gap measured by $\Delta_{19}$, the difference between the logits of
the most probable and twentieth-most probable tokens. We compare the four scores on both within-question and
cross-question targets, examine how local generation patterns influence these scores,
and study how scoring different semantic segments and reasoning positions
changes discrimination.

\section{What Does Token-Level Certainty Predict?}
\label{sec:logit-perspective}

In this section, we ask what token-level certainty tells us about a generated
response and its question. To interpret this information, we examine how
certainty relates to response correctness and question-level empirical accuracy,
and how scores vary with local token characteristics. This helps clarify what
a high certainty score indicates and how its interpretation depends on the
measurement. Unless otherwise specified, we score each response by averaging
its token scores.

\begin{figure}[!htb]
  \centering
  \includegraphics[width=\linewidth]{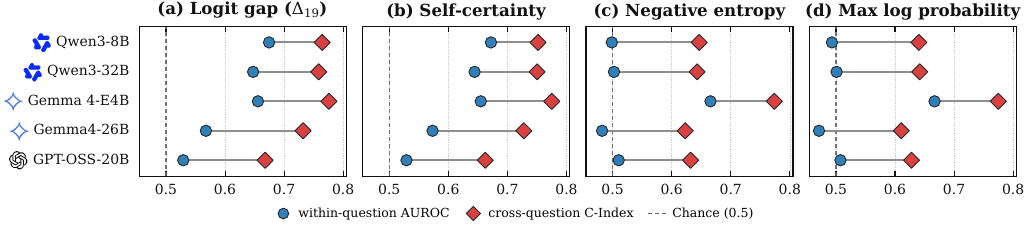}
  \setlength{\abovecaptionskip}{3pt}
  \caption{Within-question and cross-question discrimination
  across five models and four certainty metrics.
  Each panel shows one metric, with rows linking each model's within-question
  AUROC (blue circles) and cross-question C-Index (red diamonds).
  Points average results equally over AIME 2024, AIME 2025, and GPQA-Diamond.}
  \label{fig:target-comparison}
\end{figure}

\subsection{Certainty's Predictive Value Depends on the Target}
\label{subsec:target-discrimination}

Figure~\ref{fig:target-comparison} compares the four certainty metrics on both
prediction targets, with results averaged equally across the three datasets
for each model. Cross-question discrimination is higher than within-question
discrimination for every evaluated model--metric pair, although the absolute
level of discrimination varies substantially across models.
GPT-OSS-20B has near-chance within-question discrimination
under all four metrics despite retaining a clearer cross-question signal.
Gemma 4-E4B has comparatively strong discrimination on both targets.
Discrimination also varies across certainty metrics for the same model
and prediction target.
For both Qwen models and Gemma4-26B, logit gap
and self-certainty yield higher discrimination on both targets than negative
entropy and maximum log probability; the latter two give within-question AUROC values
near or below chance.
Gemma 4-E4B shows little variation across metrics.
Appendix~\ref{app:target-details} provides paired confidence intervals, compares
the two targets on a common question set, and examines sensitivity to candidate rank.

These results show that the predictive value of token-level certainty depends
on the target. In our comparisons, it distinguishes differences in empirical
accuracy across questions more strongly than correctness differences among
responses to the same question, with within-question discrimination varying
substantially across models and metrics. This distinction bears on the conditions under which
certainty can support response selection \citep{Kang2025Scalable},
likelihood-based action selection \citep{Wang2024Soft}, and reinforcement
learning with relative self-certainty among responses to the same question
\citep{Zhao2025Learning}. These uses rely on scores to identify outputs worth
selecting or reinforcing for a given input. The differences we observe across
models suggest that the reliability of these scores may depend on the model. Meanwhile, the stronger cross-question signal provides information
about which questions a model is more likely to answer correctly. Work on
allocating computation according to question difficulty \citep{Wang2024Make}
and routing questions between models \citep{Chuang2024Learning} demonstrates
the value of decisions at the question level. Our findings motivate further
study of how certainty obtained directly during generation can inform such
decisions.

\FloatBarrier
\subsection{Certainty Reflects Local Token Characteristics}
\label{sec:certainty-noise}

The preceding comparisons show that certainty's ranking ability depends on
the prediction target, model, and metric. To examine what else certainty
reflects, we now consider its association with local token characteristics.
Specifically, we ask whether token categories and positions within words
exhibit systematic differences in certainty that contribute to the scores
aggregated over a response.

\begin{figure}[!htb]
  \centering
  \includegraphics[width=\linewidth]{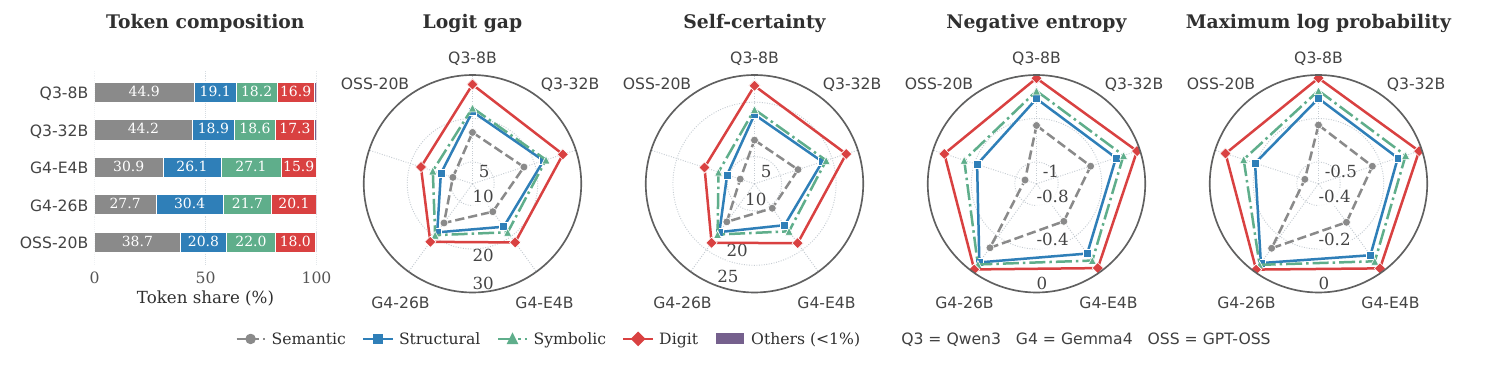}
  \setlength{\abovecaptionskip}{6pt}
  \caption{Structural, Symbolic, and Digit tokens have higher mean certainty
  than Semantic tokens across four metrics
  and five models. The left panel shows each category's share of all tokens,
  including Others (below 1\% for each model); the four radar panels show category means. Radar axes denote models; colors denote token categories.
  We average shares and means equally over the three datasets.
  The model abbreviations G4-E4B and G4-26B denote
  Gemma 4-E4B and Gemma4-26B, respectively.}
  \label{fig:token-categories}
\end{figure}

\textbf{Token Categories.} We compare four categories defined by the characters
in each token: Semantic tokens contain word-like text; Structural tokens
represent formatting or punctuation; Symbolic tokens contain mathematical
symbols or variables without digits; and Digit tokens contain digits.
Figure~\ref{fig:token-categories} shows that category shares differ and vary
across models. Structural, Symbolic, and Digit tokens have higher mean scores
than Semantic tokens across all evaluated models and metrics.
Differences in category proportions can therefore change responses' average
certainty scores. Appendix~\ref{app:surface-form-diagnostics} provides
classification details and further analyses.

\begin{figure}[!htb]
  \centering
  \includegraphics[width=\linewidth]{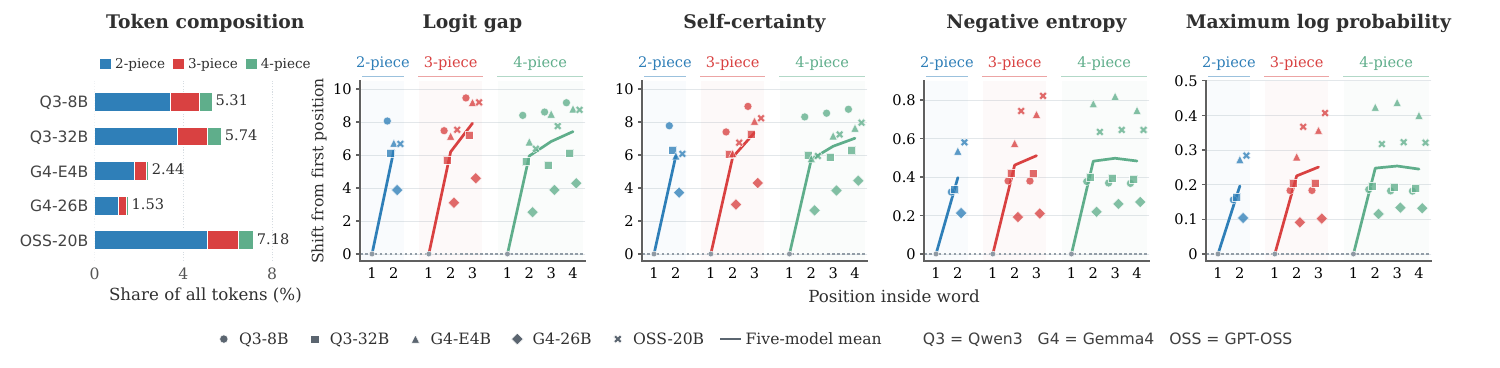}
  \setlength{\abovecaptionskip}{6pt}
  \caption{Later subwords have higher mean certainty than the first across the
  four metrics and five models shown.
  Colors distinguish words split into 2, 3, or 4 tokens. The left panel shows these detected words'
  share of all token occurrences.
  Model symbols in the four right panels show each position's
  mean minus the first-position mean for the same word length.
  Positive differences show higher certainty after the first subword
  across metrics and word lengths. G4-E4B denotes Gemma 4-E4B;
  G4-26B denotes Gemma4-26B.}
  \label{fig:subword-positions}
\end{figure}

\textbf{Positions within Words.} A word can be split into tokens called subwords \citep{clark2025doesfirsttokenentropyapproximate}.
Figure~\ref{fig:subword-positions} compares certainty at later subword positions
with the first position, grouping words by token count. Later subwords have higher mean
scores than the first across metrics and models, although the differences
need not increase at every position. Higher certainty may reflect word
completion: the first subword helps select a word, while later pieces complete
its spelling or grammatical form given the preceding pieces.
Removing later pieces of detected multi-token words and numbers, while leaving
other tokens unchanged, does not consistently improve ranking based on the
average logit gap of the retained tokens; changes vary with the prediction
target, model, and dataset (Appendix~\ref{app:mask-noisy-token}).

\FloatBarrier

\section{Where Do Effective Signals Concentrate?}
\label{sec:signal-localization}

Section~\ref{sec:logit-perspective} examined the signals in token-level certainty.
These signals contain information both relevant and irrelevant to our prediction
targets. We therefore examine how aggregating them across response regions can
improve prediction.
Prior work has aggregated certainty within response segments, as in the
trial-answer scoring used by DEER~\citep{Yang2025Dynamic} and
CGR~\citep{Nogueira2025Certainty}. Other work has aggregated certainty at
different positions in reasoning traces, including tail tokens and local
windows in DeepConf~\citep{Fu2025Deep} and prefixes in
\citet{Kim2026Reasoning}.
However, selection accuracy and compute savings also depend on candidate
responses and decision rules, so they alone cannot establish local certainty's
discriminative ability for different prediction targets
(Section~\ref{sec:ranking-and-utility}). We therefore hold the certainty metric,
generated traces, and correctness labels fixed while changing only the scored
region, comparing reasoning, summary, and final-answer segments as well as
matched-length prefix and suffix windows. By measuring discrimination for
within-question correctness and cross-question empirical accuracy across
models and tasks, we examine which regions better distinguish candidate
responses to the same question, which better indicate question difficulty,
and whether these advantages are consistent.

\subsection{Reasoning Segments}
\label{subsec:think-marked-segments}

\begin{figure}[!htb]
  \centering
  \includegraphics[width=\linewidth]{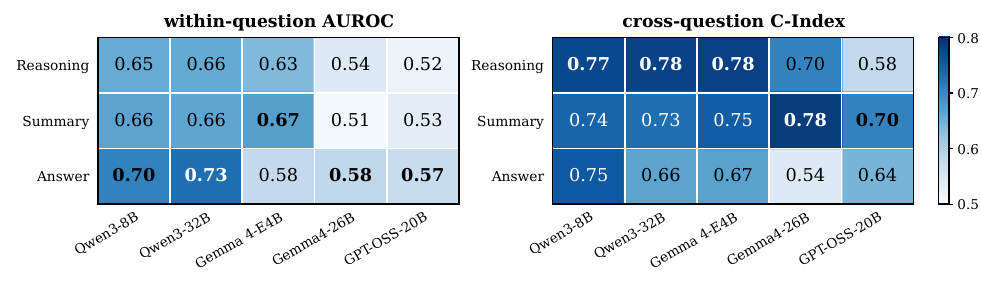}
  \caption{Segment selection changes the target captured by certainty. Cells show each model's logit-gap ranking scores for reasoning, summary, and answer tokens, averaged equally across the three datasets. Panels share a color scale; boldface marks each model's highest score per target. Answer tokens more often favor within-question ranking, whereas reasoning or summary tokens more often favor cross-question ranking.}
  \label{fig:segment-localization}
\end{figure}

We compare three regions of each trace: reasoning, summary, and final-answer
tokens. The model's output-format boundaries define the reasoning and summary
regions. In Qwen3~\citep{Yang2025Qwen3}, for example, reasoning lies between
\texttt{<think>} and \texttt{</think>}, and summary denotes the response after
the closing marker. Final-answer tokens are the extracted answer span within
that response, so the answer and summary regions overlap. We average logit gap
over each region and evaluate the resulting scores using within-question AUROC
and cross-question C-Index (Figure~\ref{fig:segment-localization}). This comparison
tests whether retaining the reasoning, the subsequent response, or only the
answer changes certainty's ability to discriminate between questions or responses.

Segment choice affects the two targets differently. With scores averaged across
datasets within each model, answer tokens give the highest within-question AUROC for
both Qwen models, Gemma4-26B, and GPT-OSS-20B; Gemma 4-E4B instead favors
the summary. The overall answer-token AUROC is $0.633$, indicating that even
the strongest average segment score provides limited correctness discrimination.
For cross-question ranking, every model favors reasoning or summary over
answer-only scoring. Restricting the score to the final answer therefore often
helps distinguish alternative responses to the same question, while losing
information useful for ranking questions by empirical accuracy. The preferred
segment therefore depends on the prediction target. Because the segments
differ in length and position, we compare temporal windows to locate each
signal along a trace.

\subsection{Temporal Localization}

\begin{figure}[!t]
  \centering
  \includegraphics[trim=0 3.060bp 0 0,clip,width=0.95\linewidth]{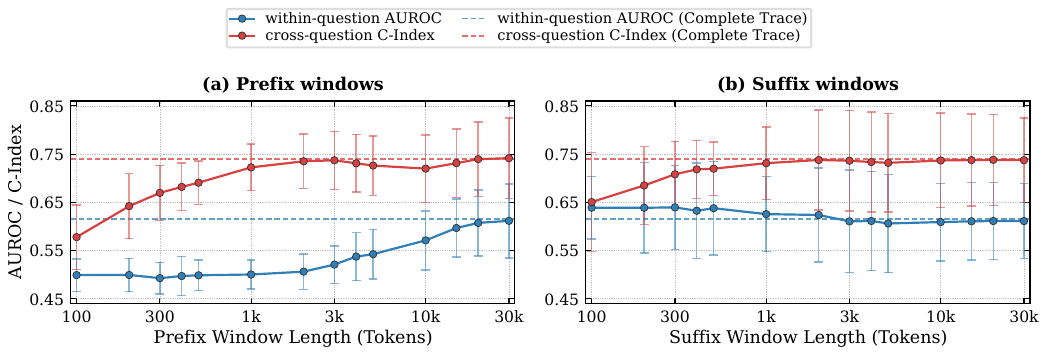}
  \caption{Temporal windows locate certainty signals within a trace. Panels (a) and (b) show absolute prefix and suffix windows, respectively. Averaging logit gap over early prefixes quickly recovers cross-question accuracy ranking, while early within-question AUROC remains near chance. Short suffix windows already provide a stronger within-question trace-correctness signal. Lines show averages across five models and three datasets; error bars show the standard deviation across model-dataset settings.}
  \label{fig:temporal-localization}
\end{figure}

We next compare prefix and suffix windows of matched token lengths to examine where certainty is informative without relying on model-specific segment boundaries. For each window length, we average the logit gap over the selected tokens and evaluate within-question AUROC and cross-question C-Index (Figure~\ref{fig:temporal-localization}). Traces shorter than the specified window are scored in full. Appendix~\ref{app:signal-localization-details} examines whether the same trends hold for windows defined as a fraction of trace length.

For cross-question ranking, the average C-Index rises rapidly as either prefix or suffix windows expand and largely plateaus at around 1,000 tokens. Both prefixes and suffixes therefore contain information for ranking questions by empirical accuracy. In particular, prefix scores provide an informative ranking early in generation. Thus, even before a response is complete, its opening tokens can provide an initial indication of which questions the model is more likely to answer correctly.

For within-question trace selection, the temporal asymmetry is sharper. Short prefixes have little discriminative power, with AUROC close to chance; the signal strengthens as the observed window moves toward the end of the trace. Suffix windows, especially short suffixes, are already more informative than prefix windows of the same length, and some suffix aggregations exceed the full-trace average. This suggests that trace-level correctness evidence is more concentrated near the final answer, where the model commits to and resolves a specific sampled solution. Comparisons across all four certainty metrics likewise show informative prefixes for cross-question ranking and higher average within-question AUROC for suffixes than for full traces (Appendix~\ref{app:temporal-metric-comparisons}). Together, these findings motivate the division of operations in TACS (Section~\ref{sec:test-time-compute}). Earlier or wider-window certainty is better matched to cross-question budget allocation, while late-window certainty is better suited to post-generation candidate filtering or weighting.

\Needspace*{4\baselineskip}
\section{How Can Certainty Signals Guide Test-Time Compute?}
\label{sec:test-time-compute}

Our analyses show that token-level certainty's discriminative power varies with the prediction target, scoring metric, and aggregation region. We translate these findings into Target-Aligned Certainty Scaling (TACS), a test-time scaling (TTS) framework that selects a scoring metric and matches signals from different regions to the compute decisions they support. We use logit gap because it performs well in the preceding ranking comparisons, is inexpensive to compute directly from logit differences, and preserves rankings under uniform temperature scaling. Guided by the differences in strength and localization between cross-question and within-question signals, TACS uses early certainty to allocate sampling budgets and late certainty to weight candidate traces' votes. We first test each decision's contribution independently, then evaluate the combined framework's compute--accuracy trade-off.

\paragraph{Early allocation.}

Our analyses show that certainty better distinguishes empirical performance across questions and that short prefixes already capture a strong cross-question signal (Section~\ref{sec:signal-localization}). We can therefore control each question's sample count from prefix scores alone. A trace counts toward the sampling target if its mean logit gap over the first 1,000 tokens is at least $\theta_m$, a model-specific threshold that is calibrated on a separate dataset and held fixed across the evaluation datasets. Appendix~\ref{app:tts-protocol} reports the model-specific thresholds. We fix $N=32$ throughout this section, sampling until 32 traces meet the threshold or we observe 256 traces. This shared target yields different sample counts across questions: those that more often produce high-certainty prefixes reach the target with fewer samples, whereas others continue sampling. For all five models, the hardest question group receives more samples on average than the easiest group (Appendix~\ref{app:sampling-allocation-stopping}). We compare prefix allocation with fixed 256-trace sampling across voting methods, using the same candidate set for all methods under each strategy. In Table~\ref{tab:tts-component-ablation}, prefix allocation reduces generated-token cost by $82.44\%$, while overall accuracy decreases by only 0.17--0.33 percentage points (pp) relative to fixed-budget sampling when both sampling strategies use the same voting method.

\begin{table}[!htbp]
  \centering
  \small
  \caption{Answer aggregation under fixed-budget sampling and early allocation.
  We compare majority voting with rank-weighted voting using certainty computed over the full trace
  or the final 1,000 or 500 tokens. Within each sampling run, all voting methods share candidate responses
  and thus generated-token cost. Rank-weighted voting based on the final 500 tokens has the highest
  overall accuracy among the four methods under both sampling conditions. With early allocation,
  it reaches 79.54\% accuracy at $0.176\times$ the token cost of fixed 256-trace sampling.}
  \label{tab:tts-component-ablation}
  \setlength{\tabcolsep}{0pt}
  \renewcommand{\arraystretch}{1.05}
  \begin{tabular}{@{}p{0.16\linewidth}*{10}{>{\centering\arraybackslash}p{0.084\linewidth}}@{}}
    \toprule
    & \multicolumn{5}{c}{\textbf{Fixed Budget}}
    & \multicolumn{5}{c}{\textbf{Early allocation}} \\
    \cmidrule(lr){2-6}\cmidrule(l){7-11}
    & \multicolumn{4}{c}{Accuracy (\%) $\uparrow$}
    & \multicolumn{1}{c}{Cost $\downarrow$}
    & \multicolumn{4}{c}{Accuracy (\%) $\uparrow$}
    & \multicolumn{1}{c}{Cost $\downarrow$} \\
    \cmidrule(lr){2-5}\cmidrule(lr){7-10}
    \textbf{Model} & \makebox[0pt][c]{\footnotesize Majority} & \makebox[0pt][c]{\footnotesize Full} & \makebox[0pt][c]{\footnotesize Suffix 1k} & \makebox[0pt][c]{\footnotesize Suffix 500} &
    & \makebox[0pt][c]{\footnotesize Majority} & \makebox[0pt][c]{\footnotesize Full} & \makebox[0pt][c]{\footnotesize Suffix 1k} & \makebox[0pt][c]{\footnotesize Suffix 500} & \\
    \midrule
    \qwenicon\hspace{0.2em}Qwen3-8B & 75.43 & 76.15 & 76.36 & \textbf{76.45} & 1.000 & 74.96 & 75.73 & 76.21 & \textbf{76.33} & 0.142 \\
    \qwenicon\hspace{0.2em}Qwen3-32B & 79.63 & 81.11 & 80.78 & \textbf{81.31} & 1.000 & 80.68 & 81.38 & 81.37 & \textbf{81.75} & 0.145 \\
    \gemmaicon\hspace{0.2em}Gemma4-E4B & 62.11 & 63.72 & 63.64 & \textbf{64.38} & 1.000 & 61.22 & 62.92 & 63.10 & \textbf{63.80} & 0.186 \\
    \gemmaicon\hspace{0.2em}Gemma4-26B & 91.39 & \textbf{91.51} & 91.37 & 91.49 & 1.000 & 90.87 & \textbf{91.13} & 90.71 & 90.86 & 0.192 \\
    \openaiicon\hspace{0.2em}GPT-OSS-20B & \textbf{84.98} & 84.78 & 84.97 & 84.94 & 1.000 & 84.46 & 84.48 & 84.87 & \textbf{84.96} & 0.214 \\
    \midrule
    \textbf{Overall} & 78.71 & 79.45 & 79.42 & \textbf{79.71} & 1.000 & 78.44 & 79.13 & 79.25 & \textbf{79.54} & 0.176 \\
    \bottomrule
  \end{tabular}
\end{table}

\paragraph{Late voting.}

Within-question information is more concentrated near the end of a trace (Section~\ref{sec:signal-localization}), motivating suffix certainty as an auxiliary ranking signal for voting. We rank sampled responses with valid answers by their mean logit gap over the final 500 tokens. Each response receives a vote weight equal to its ascending certainty rank, so higher-certainty responses receive larger weights. Responses below the prefix threshold also participate in voting, and the answer with the largest total vote weight wins. This leaves sampling cost unchanged. With prefix allocation and fixed 256-trace sampling, rank-weighted voting based on certainty over the final 500 tokens improves overall accuracy over majority voting by $1.10$ and $1.01$ pp, respectively. Four of five models improve under each condition (Table~\ref{tab:tts-component-ablation}). Compared with full-trace rank-weighted voting, the gains are $0.41$ and $0.26$ pp, respectively, although Gemma4-26B shows the reverse ordering under both conditions. These results echo our earlier observations: the relatively weak within-question signal is stronger near the end of a trace, while gains in answer aggregation remain model-dependent.

\paragraph{Main results.}

We evaluate TACS against a fixed-budget self-consistency baseline, $\mathrm{cons}@256$, which uses majority voting over 256 traces. As shown in Table~\ref{tab:tts-component-ablation}, TACS reaches $79.54\%$ accuracy at $0.176\times$ the token cost of $\mathrm{cons}@256$. Its accuracy exceeds fixed-256 majority voting by $0.83$ pp and is $0.17$ pp below fixed-256 sampling with the same voting rule based on certainty over the final 500 tokens. Appendix~\ref{app:baseline-comparisons} provides additional comparisons with fixed-budget and adaptive baselines. TACS thus uses early cross-question information for sampling decisions and late within-question information for vote weighting, selecting scoring regions by decision target.

\section{Conclusion}

Our experiments show that token-level certainty generally distinguishes questions by empirical accuracy more effectively than it distinguishes correct from incorrect responses to the same question. Its predictive value varies across models, metrics, and response regions, while its scores also reflect local token characteristics. Question-level ranking information emerges early, whereas within-question correctness information is more concentrated near the end. TACS uses this distinction to allocate samples with prefix certainty and weight answer votes with suffix certainty. In our experiments, TACS reduces generated-token cost by 82.4\% while increasing overall accuracy from 78.71\% to 79.54\% relative to majority voting over 256 responses. These findings support matching the scoring region to the prediction target and evaluating certainty both as a predictor and as part of the decision rule that uses it. We believe these insights can guide more effective uses of certainty across LLM training and inference.

\clearpage
\bibliographystyle{abbrvnat}
\bibliography{references}

\appendix

\clearpage
\raggedbottom
\renewcommand{\floatpagefraction}{0.8}

\section{Details for Section 4.1}
\label{app:target-details}

The analyses in Appendices~\ref{app:target-comparison}--\ref{app:common-question-set}
use the same full-trace scores as
Figure~\ref{fig:target-comparison}: logit gap $\Delta_{19}$, self-certainty,
negative entropy, and maximum log probability, computed from the full-softmax
probabilities of the 20 most probable tokens, without renormalization.
For each complete response, we average its token scores.
All four metrics use the same set of responses; sample sizes are given in
Appendix~\ref{app:diagnostic-samples}.

\subsection{Discrimination across Models and Datasets}
\label{app:target-comparison}
Tables~\ref{tab:target-comparison-by-model} and
\ref{tab:target-comparison-by-dataset} report within-question AUROC and
cross-question C-Index as defined in Section~\ref{sec:ranking-and-utility}.
We compute statistics within each model--dataset setting and report
equal-weight averages by model, by dataset, and overall.
Cross-question C-Index exceeds within-question AUROC in all 60
model--dataset--metric combinations. Paired 95\% intervals for the difference
are above zero in 34 of the 60 combinations and in all model-level,
dataset-level, and overall averages.
Logit gap and self-certainty have higher within-question discrimination on
the two AIME datasets than on GPQA-Diamond, whereas negative entropy and
maximum log probability show the opposite ordering. The cross-question and
within-question statistics concern distinct prediction targets.

\begin{table}[!htbp]
  \centering
  \footnotesize
  \newcommand{\sctci}[3]{\begingroup\renewcommand{\arraystretch}{0.88}%
    \begin{tabular}{@{}c@{}}#1\\[-1.3pt]{[#2,\,#3]}\end{tabular}\endgroup}
  \caption{Full-trace within-question AUROC and cross-question C-Index by model. Cross-question discrimination is higher for every model--metric average.
  Model rows average three datasets equally; Overall averages all 15 settings equally.
  Evaluation follows Section~\ref{sec:ranking-and-utility}.
  Gemma4-4B denotes Gemma 4-E4B.
  Blue ``Yes'' denotes a positive difference, not statistical significance.
  Brackets give pointwise 95\% paired question-bootstrap confidence intervals
  (Appendix~\ref{app:diagnostic-samples}).}
  \label{tab:target-comparison-by-model}
  \setlength{\tabcolsep}{2.4pt}
  \renewcommand{\arraystretch}{1.08}
  \begin{tabular*}{\linewidth}{@{\extracolsep{\fill}}lcccc@{}}
    \toprule
    \textbf{Model}
    & \shortstack{\textbf{within-question}\\AUROC (W)}
    & \shortstack{\textbf{cross-question}\\C-Index (C)}
    & \shortstack{\textbf{C $-$ W}\\95\% CI}
    & \shortstack{\textbf{cross-question}\\\textbf{$>$ within-question?}} \\
    \midrule
    \multicolumn{5}{@{}l@{}}{\textbf{A. Logit gap ($\Delta_{19}$)}} \\
    \addlinespace[2pt]
\qwenicon\hspace{0.20em}Qwen3-8B & 0.674 & 0.764 & \sctci{+0.0899}{0.0396}{0.1424} & \cellcolor{rankStable}Yes \\
    \addlinespace[0.5pt]
\qwenicon\hspace{0.20em}Qwen3-32B & 0.647 & 0.758 & \sctci{+0.1109}{0.0549}{0.1631} & \cellcolor{rankStable}Yes \\
    \addlinespace[0.5pt]
\gemmaicon\hspace{0.20em}Gemma4-4B & 0.656 & 0.776 & \sctci{+0.1202}{0.0565}{0.1860} & \cellcolor{rankStable}Yes \\
    \addlinespace[0.5pt]
\gemmaicon\hspace{0.20em}Gemma4-26B & 0.568 & 0.732 & \sctci{+0.1645}{0.0603}{0.2741} & \cellcolor{rankStable}Yes \\
    \addlinespace[0.5pt]
\openaiicon\hspace{0.20em}GPT-OSS-20B & 0.529 & 0.668 & \sctci{+0.1385}{0.0686}{0.2041} & \cellcolor{rankStable}Yes \\
    \addlinespace[0.5pt]
    \cmidrule(l){1-5}
    \textbf{Overall} & 0.615 & 0.740 & \sctci{+0.1248}{0.0829}{0.1677} & \cellcolor{rankStable}Yes \\
    \addlinespace[0.5pt]
    \midrule
    \multicolumn{5}{@{}l@{}}{\textbf{B. Self-certainty}} \\
    \addlinespace[2pt]
\qwenicon\hspace{0.20em}Qwen3-8B & 0.672 & 0.752 & \sctci{+0.0798}{0.0244}{0.1354} & \cellcolor{rankStable}Yes \\
    \addlinespace[0.5pt]
\qwenicon\hspace{0.20em}Qwen3-32B & 0.644 & 0.750 & \sctci{+0.1059}{0.0492}{0.1589} & \cellcolor{rankStable}Yes \\
    \addlinespace[0.5pt]
\gemmaicon\hspace{0.20em}Gemma4-4B & 0.655 & 0.775 & \sctci{+0.1202}{0.0553}{0.1867} & \cellcolor{rankStable}Yes \\
    \addlinespace[0.5pt]
\gemmaicon\hspace{0.20em}Gemma4-26B & 0.573 & 0.728 & \sctci{+0.1545}{0.0523}{0.2660} & \cellcolor{rankStable}Yes \\
    \addlinespace[0.5pt]
\openaiicon\hspace{0.20em}GPT-OSS-20B & 0.529 & 0.662 & \sctci{+0.1332}{0.0634}{0.1993} & \cellcolor{rankStable}Yes \\
    \addlinespace[0.5pt]
    \cmidrule(l){1-5}
    \textbf{Overall} & 0.615 & 0.733 & \sctci{+0.1187}{0.0755}{0.1613} & \cellcolor{rankStable}Yes \\
    \addlinespace[0.5pt]
    \midrule
    \multicolumn{5}{@{}l@{}}{\textbf{C. Negative entropy}} \\
    \addlinespace[2pt]
\qwenicon\hspace{0.20em}Qwen3-8B & 0.499 & 0.647 & \sctci{+0.1479}{0.0693}{0.2248} & \cellcolor{rankStable}Yes \\
    \addlinespace[0.5pt]
\qwenicon\hspace{0.20em}Qwen3-32B & 0.502 & 0.643 & \sctci{+0.1407}{0.0726}{0.2036} & \cellcolor{rankStable}Yes \\
    \addlinespace[0.5pt]
\gemmaicon\hspace{0.20em}Gemma4-4B & 0.666 & 0.774 & \sctci{+0.1080}{0.0407}{0.1814} & \cellcolor{rankStable}Yes \\
    \addlinespace[0.5pt]
\gemmaicon\hspace{0.20em}Gemma4-26B & 0.482 & 0.623 & \sctci{+0.1406}{0.0349}{0.2476} & \cellcolor{rankStable}Yes \\
    \addlinespace[0.5pt]
\openaiicon\hspace{0.20em}GPT-OSS-20B & 0.510 & 0.632 & \sctci{+0.1218}{0.0472}{0.1919} & \cellcolor{rankStable}Yes \\
    \addlinespace[0.5pt]
    \cmidrule(l){1-5}
    \textbf{Overall} & 0.532 & 0.664 & \sctci{+0.1318}{0.0888}{0.1758} & \cellcolor{rankStable}Yes \\
    \addlinespace[0.5pt]
    \midrule
    \multicolumn{5}{@{}l@{}}{\textbf{D. Maximum log probability}} \\
    \addlinespace[2pt]
\qwenicon\hspace{0.20em}Qwen3-8B & 0.493 & 0.640 & \sctci{+0.1472}{0.0669}{0.2234} & \cellcolor{rankStable}Yes \\
    \addlinespace[0.5pt]
\qwenicon\hspace{0.20em}Qwen3-32B & 0.501 & 0.642 & \sctci{+0.1407}{0.0733}{0.2039} & \cellcolor{rankStable}Yes \\
    \addlinespace[0.5pt]
\gemmaicon\hspace{0.20em}Gemma4-4B & 0.667 & 0.775 & \sctci{+0.1079}{0.0421}{0.1800} & \cellcolor{rankStable}Yes \\
    \addlinespace[0.5pt]
\gemmaicon\hspace{0.20em}Gemma4-26B & 0.471 & 0.610 & \sctci{+0.1393}{0.0355}{0.2465} & \cellcolor{rankStable}Yes \\
    \addlinespace[0.5pt]
\openaiicon\hspace{0.20em}GPT-OSS-20B & 0.508 & 0.628 & \sctci{+0.1203}{0.0458}{0.1897} & \cellcolor{rankStable}Yes \\
    \addlinespace[0.5pt]
    \cmidrule(l){1-5}
    \textbf{Overall} & 0.528 & 0.659 & \sctci{+0.1311}{0.0886}{0.1744} & \cellcolor{rankStable}Yes \\
    \addlinespace[0.5pt]
    \bottomrule
  \end{tabular*}
\end{table}

\begin{table}[!htbp]
  \centering
  \footnotesize
  \newcommand{\sctci}[3]{\begingroup\renewcommand{\arraystretch}{0.88}%
    \begin{tabular}{@{}c@{}}#1\\[-1.3pt]{[#2,\,#3]}\end{tabular}\endgroup}
  \caption{Full-trace within-question AUROC and cross-question C-Index by dataset. Cross-question discrimination is higher for every dataset--metric average.
  Dataset rows average five models equally; Overall averages all 15 settings equally.
  Evaluation follows Section~\ref{sec:ranking-and-utility}.
  Blue ``Yes'' denotes a positive difference, not statistical significance.
  Brackets give pointwise 95\% paired question-bootstrap intervals
  (Appendix~\ref{app:diagnostic-samples}).}
  \label{tab:target-comparison-by-dataset}
  \setlength{\tabcolsep}{2.4pt}
  \renewcommand{\arraystretch}{1.08}
  \begin{tabular*}{\linewidth}{@{\extracolsep{\fill}}lcccc@{}}
    \toprule
    \textbf{Dataset}
    & \shortstack{\textbf{within-question}\\AUROC (W)}
    & \shortstack{\textbf{cross-question}\\C-Index (C)}
    & \shortstack{\textbf{C $-$ W}\\95\% CI}
    & \shortstack{\textbf{cross-question}\\\textbf{$>$ within-question?}} \\
    \midrule
    \multicolumn{5}{@{}l@{}}{\textbf{A. Logit gap ($\Delta_{19}$)}} \\
    \addlinespace[2pt]
AIME 2024 & 0.624 & 0.759 & \sctci{+0.1348}{0.0548}{0.2138} & \cellcolor{rankStable}Yes \\
    \addlinespace[0.5pt]
AIME 2025 & 0.637 & 0.783 & \sctci{+0.1458}{0.0500}{0.2373} & \cellcolor{rankStable}Yes \\
    \addlinespace[0.5pt]
GPQA-Diamond & 0.583 & 0.677 & \sctci{+0.0938}{0.0550}{0.1309} & \cellcolor{rankStable}Yes \\
    \addlinespace[0.5pt]
    \cmidrule(l){1-5}
    \textbf{Overall} & 0.615 & 0.740 & \sctci{+0.1248}{0.0829}{0.1677} & \cellcolor{rankStable}Yes \\
    \addlinespace[0.5pt]
    \midrule
    \multicolumn{5}{@{}l@{}}{\textbf{B. Self-certainty}} \\
    \addlinespace[2pt]
AIME 2024 & 0.623 & 0.749 & \sctci{+0.1265}{0.0464}{0.2033} & \cellcolor{rankStable}Yes \\
    \addlinespace[0.5pt]
AIME 2025 & 0.638 & 0.778 & \sctci{+0.1396}{0.0429}{0.2322} & \cellcolor{rankStable}Yes \\
    \addlinespace[0.5pt]
GPQA-Diamond & 0.583 & 0.673 & \sctci{+0.0900}{0.0508}{0.1282} & \cellcolor{rankStable}Yes \\
    \addlinespace[0.5pt]
    \cmidrule(l){1-5}
    \textbf{Overall} & 0.615 & 0.733 & \sctci{+0.1187}{0.0755}{0.1613} & \cellcolor{rankStable}Yes \\
    \addlinespace[0.5pt]
    \midrule
    \multicolumn{5}{@{}l@{}}{\textbf{C. Negative entropy}} \\
    \addlinespace[2pt]
AIME 2024 & 0.522 & 0.692 & \sctci{+0.1701}{0.0922}{0.2489} & \cellcolor{rankStable}Yes \\
    \addlinespace[0.5pt]
AIME 2025 & 0.526 & 0.657 & \sctci{+0.1317}{0.0402}{0.2264} & \cellcolor{rankStable}Yes \\
    \addlinespace[0.5pt]
GPQA-Diamond & 0.547 & 0.641 & \sctci{+0.0936}{0.0465}{0.1365} & \cellcolor{rankStable}Yes \\
    \addlinespace[0.5pt]
    \cmidrule(l){1-5}
    \textbf{Overall} & 0.532 & 0.664 & \sctci{+0.1318}{0.0888}{0.1758} & \cellcolor{rankStable}Yes \\
    \addlinespace[0.5pt]
    \midrule
    \multicolumn{5}{@{}l@{}}{\textbf{D. Maximum log probability}} \\
    \addlinespace[2pt]
AIME 2024 & 0.518 & 0.685 & \sctci{+0.1671}{0.0887}{0.2495} & \cellcolor{rankStable}Yes \\
    \addlinespace[0.5pt]
AIME 2025 & 0.520 & 0.652 & \sctci{+0.1326}{0.0419}{0.2283} & \cellcolor{rankStable}Yes \\
    \addlinespace[0.5pt]
GPQA-Diamond & 0.546 & 0.639 & \sctci{+0.0935}{0.0466}{0.1361} & \cellcolor{rankStable}Yes \\
    \addlinespace[0.5pt]
    \cmidrule(l){1-5}
    \textbf{Overall} & 0.528 & 0.659 & \sctci{+0.1311}{0.0886}{0.1744} & \cellcolor{rankStable}Yes \\
    \addlinespace[0.5pt]
    \bottomrule
  \end{tabular*}
\end{table}

\subsection{Correctness Comparisons within and across Questions}
\label{app:correctness-pairs}
To hold the prediction outcome fixed, we compare correctness concordance
for same-question and different-question trace pairs, denoted by
$A_{\mathrm{same}}$ and $A_{\mathrm{diff}}$. Pairs are equally weighted
within each group, with half credit for score ties. If the corresponding
correct--incorrect pair counts are $N_{\mathrm{same}}$ and $N_{\mathrm{diff}}$,
pooled AUROC within a model--dataset setting is
\begin{equation}
\label{eq:pooled-correctness-decomposition}
\mathrm{AUROC}_{\mathrm{pooled}}
=\frac{N_{\mathrm{same}}A_{\mathrm{same}}+
N_{\mathrm{diff}}A_{\mathrm{diff}}}
{N_{\mathrm{same}}+N_{\mathrm{diff}}}.
\end{equation}
Unlike the question-equal within-question AUROC, $A_{\mathrm{same}}$ weights
each question by its number of correct--incorrect pairs.
$A_{\mathrm{diff}}$ compares individual trace scores against correctness
labels; unlike cross-question C-Index, it does not compare mean question
scores against empirical accuracy.

Tables~\ref{tab:correctness-pair-by-model} and
\ref{tab:correctness-pair-by-dataset} report concordance for both groups and pooled AUROC.
Different-question concordance exceeds same-question concordance in all
60 model--dataset--metric combinations. Paired 95\% intervals for this
difference are above zero in 47 combinations. For logit gap, the overall values of same-question concordance,
different-question concordance, and pooled AUROC are $0.583$, $0.801$, and $0.799$, respectively. Different-question
pairs account for $99.05\%$ of correct--incorrect pairs on average across
settings, so pooled AUROC can mask weak within-question discrimination.

\begin{table}[!htbp]
  \centering
  \footnotesize
  \newcommand{\sctci}[3]{\begingroup\renewcommand{\arraystretch}{0.80}%
    \begin{tabular}{@{}c@{}}#1\\[-1.3pt]{[#2,\,#3]}\end{tabular}\endgroup}
  \caption{Correctness discrimination by model under equal trace-pair weighting. Different-question concordance exceeds same-question concordance for every model--metric average.
  Model rows average three datasets equally; Overall averages all 15 settings equally.
  $A_{\mathrm{same}}$ and $A_{\mathrm{diff}}$ measure correctness concordance for
  equally weighted trace pairs from the same and different questions, respectively.
  Pooled AUROC is computed within each setting
  using Equation~\ref{eq:pooled-correctness-decomposition} before averaging.
  Blue ``Yes'' denotes a positive difference, not statistical significance.
  Brackets give pointwise 95\% paired question-bootstrap intervals
  (Appendix~\ref{app:diagnostic-samples}).}
  \label{tab:correctness-pair-by-model}
  \setlength{\tabcolsep}{3.2pt}
  \renewcommand{\arraystretch}{1.08}
  \begin{tabular*}{\linewidth}{@{\extracolsep{\fill}}lccccc@{}}
    \toprule
    \textbf{Model}
    & \shortstack{\textbf{Same-question}\\concordance ($A_{\mathrm{same}}$)}
    & \shortstack{\textbf{Different-question}\\concordance ($A_{\mathrm{diff}}$)}
    & \shortstack{\textbf{Pooled}\\AUROC}
    & \shortstack{\textbf{Difference}\\$A_{\mathrm{diff}}-A_{\mathrm{same}}$ [95\% CI]}
    & \shortstack{\textbf{Different $>$ Same?}\\Point estimate} \\
    \midrule
    \multicolumn{6}{@{}l@{}}{\textbf{A. Logit gap ($\Delta_{19}$)}} \\
    \addlinespace[2pt]
\qwenicon\hspace{0.20em}Qwen3-8B & 0.633 & 0.869 & 0.866 & \sctci{+0.2360}{0.1674}{0.3033} & \cellcolor{rankStable}Yes \\
    \addlinespace[0.5pt]
\qwenicon\hspace{0.20em}Qwen3-32B & 0.642 & 0.848 & 0.846 & \sctci{+0.2061}{0.1406}{0.2757} & \cellcolor{rankStable}Yes \\
    \addlinespace[0.5pt]
\gemmaicon\hspace{0.20em}Gemma4-4B & 0.619 & 0.839 & 0.836 & \sctci{+0.2195}{0.1319}{0.3088} & \cellcolor{rankStable}Yes \\
    \addlinespace[0.5pt]
\gemmaicon\hspace{0.20em}Gemma4-26B & 0.498 & 0.766 & 0.763 & \sctci{+0.2682}{0.1169}{0.3454} & \cellcolor{rankStable}Yes \\
    \addlinespace[0.5pt]
\openaiicon\hspace{0.20em}GPT-OSS-20B & 0.526 & 0.686 & 0.684 & \sctci{+0.1601}{0.0887}{0.2390} & \cellcolor{rankStable}Yes \\
    \addlinespace[0.5pt]
    \cmidrule(l){1-6}
    \textbf{Overall} & 0.583 & 0.801 & 0.799 & \sctci{+0.2180}{0.1584}{0.2638} & \cellcolor{rankStable}Yes \\
    \addlinespace[0.5pt]
    \midrule
    \multicolumn{6}{@{}l@{}}{\textbf{B. Self-certainty}} \\
    \addlinespace[2pt]
\qwenicon\hspace{0.20em}Qwen3-8B & 0.632 & 0.865 & 0.862 & \sctci{+0.2326}{0.1638}{0.3015} & \cellcolor{rankStable}Yes \\
    \addlinespace[0.5pt]
\qwenicon\hspace{0.20em}Qwen3-32B & 0.642 & 0.846 & 0.844 & \sctci{+0.2047}{0.1398}{0.2743} & \cellcolor{rankStable}Yes \\
    \addlinespace[0.5pt]
\gemmaicon\hspace{0.20em}Gemma4-4B & 0.618 & 0.838 & 0.835 & \sctci{+0.2195}{0.1314}{0.3084} & \cellcolor{rankStable}Yes \\
    \addlinespace[0.5pt]
\gemmaicon\hspace{0.20em}Gemma4-26B & 0.501 & 0.763 & 0.760 & \sctci{+0.2627}{0.1102}{0.3399} & \cellcolor{rankStable}Yes \\
    \addlinespace[0.5pt]
\openaiicon\hspace{0.20em}GPT-OSS-20B & 0.526 & 0.682 & 0.680 & \sctci{+0.1563}{0.0836}{0.2347} & \cellcolor{rankStable}Yes \\
    \addlinespace[0.5pt]
    \cmidrule(l){1-6}
    \textbf{Overall} & 0.584 & 0.799 & 0.797 & \sctci{+0.2152}{0.1554}{0.2614} & \cellcolor{rankStable}Yes \\
    \addlinespace[0.5pt]
    \midrule
    \multicolumn{6}{@{}l@{}}{\textbf{C. Negative entropy}} \\
    \addlinespace[2pt]
\qwenicon\hspace{0.20em}Qwen3-8B & 0.482 & 0.739 & 0.736 & \sctci{+0.2567}{0.1802}{0.3270} & \cellcolor{rankStable}Yes \\
    \addlinespace[0.5pt]
\qwenicon\hspace{0.20em}Qwen3-32B & 0.509 & 0.739 & 0.737 & \sctci{+0.2303}{0.1683}{0.3004} & \cellcolor{rankStable}Yes \\
    \addlinespace[0.5pt]
\gemmaicon\hspace{0.20em}Gemma4-4B & 0.622 & 0.836 & 0.833 & \sctci{+0.2140}{0.1257}{0.3055} & \cellcolor{rankStable}Yes \\
    \addlinespace[0.5pt]
\gemmaicon\hspace{0.20em}Gemma4-26B & 0.422 & 0.590 & 0.588 & \sctci{+0.1673}{0.0414}{0.2428} & \cellcolor{rankStable}Yes \\
    \addlinespace[0.5pt]
\openaiicon\hspace{0.20em}GPT-OSS-20B & 0.513 & 0.640 & 0.639 & \sctci{+0.1279}{0.0520}{0.2081} & \cellcolor{rankStable}Yes \\
    \addlinespace[0.5pt]
    \cmidrule(l){1-6}
    \textbf{Overall} & 0.509 & 0.709 & 0.707 & \sctci{+0.1992}{0.1479}{0.2433} & \cellcolor{rankStable}Yes \\
    \addlinespace[0.5pt]
    \midrule
    \multicolumn{6}{@{}l@{}}{\textbf{D. Maximum log probability}} \\
    \addlinespace[2pt]
\qwenicon\hspace{0.20em}Qwen3-8B & 0.479 & 0.730 & 0.728 & \sctci{+0.2514}{0.1755}{0.3220} & \cellcolor{rankStable}Yes \\
    \addlinespace[0.5pt]
\qwenicon\hspace{0.20em}Qwen3-32B & 0.505 & 0.733 & 0.731 & \sctci{+0.2276}{0.1648}{0.2974} & \cellcolor{rankStable}Yes \\
    \addlinespace[0.5pt]
\gemmaicon\hspace{0.20em}Gemma4-4B & 0.624 & 0.837 & 0.835 & \sctci{+0.2124}{0.1245}{0.3037} & \cellcolor{rankStable}Yes \\
    \addlinespace[0.5pt]
\gemmaicon\hspace{0.20em}Gemma4-26B & 0.416 & 0.564 & 0.562 & \sctci{+0.1483}{0.0275}{0.2241} & \cellcolor{rankStable}Yes \\
    \addlinespace[0.5pt]
\openaiicon\hspace{0.20em}GPT-OSS-20B & 0.511 & 0.633 & 0.631 & \sctci{+0.1223}{0.0454}{0.2025} & \cellcolor{rankStable}Yes \\
    \addlinespace[0.5pt]
    \cmidrule(l){1-6}
    \textbf{Overall} & 0.507 & 0.699 & 0.697 & \sctci{+0.1924}{0.1422}{0.2367} & \cellcolor{rankStable}Yes \\
    \addlinespace[0.5pt]
    \bottomrule
  \end{tabular*}
\end{table}

\begin{table}[!htbp]
  \centering
  \footnotesize
  \newcommand{\sctci}[3]{\begingroup\renewcommand{\arraystretch}{0.88}%
    \begin{tabular}{@{}c@{}}#1\\[-1.3pt]{[#2,\,#3]}\end{tabular}\endgroup}
  \caption{Correctness discrimination by dataset under equal trace-pair weighting. Different-question concordance exceeds same-question concordance for every dataset--metric average.
  Dataset rows average five models equally; Overall averages all 15 settings equally.
  $A_{\mathrm{same}}$ and $A_{\mathrm{diff}}$ measure correctness concordance for
  equally weighted trace pairs from the same and different questions, respectively.
  Pooled AUROC is computed within each setting
  using Equation~\ref{eq:pooled-correctness-decomposition} before averaging.
  Blue ``Yes'' denotes a positive difference, not statistical significance.
  Brackets give pointwise 95\% paired question-bootstrap intervals
  (Appendix~\ref{app:diagnostic-samples}).}
  \label{tab:correctness-pair-by-dataset}
  \setlength{\tabcolsep}{3.2pt}
  \renewcommand{\arraystretch}{1.08}
  \begin{tabular*}{\linewidth}{@{\extracolsep{\fill}}lccccc@{}}
    \toprule
    \textbf{Dataset}
    & \shortstack{\textbf{Same-question}\\concordance ($A_{\mathrm{same}}$)}
    & \shortstack{\textbf{Different-question}\\concordance ($A_{\mathrm{diff}}$)}
    & \shortstack{\textbf{Pooled}\\AUROC}
    & \shortstack{\textbf{Difference}\\$A_{\mathrm{diff}}-A_{\mathrm{same}}$ [95\% CI]}
    & \shortstack{\textbf{Different $>$ Same?}\\Point estimate} \\
    \midrule
    \multicolumn{6}{@{}l@{}}{\textbf{A. Logit gap ($\Delta_{19}$)}} \\
    \addlinespace[2pt]
AIME 2024 & 0.598 & 0.854 & 0.852 & \sctci{+0.2563}{0.1435}{0.3337} & \cellcolor{rankStable}Yes \\
    \addlinespace[0.5pt]
AIME 2025 & 0.587 & 0.837 & 0.833 & \sctci{+0.2508}{0.1208}{0.3512} & \cellcolor{rankStable}Yes \\
    \addlinespace[0.5pt]
GPQA-Diamond & 0.565 & 0.712 & 0.712 & \sctci{+0.1468}{0.0925}{0.1956} & \cellcolor{rankStable}Yes \\
    \addlinespace[0.5pt]
    \cmidrule(l){1-6}
    \textbf{Overall} & 0.583 & 0.801 & 0.799 & \sctci{+0.2180}{0.1584}{0.2638} & \cellcolor{rankStable}Yes \\
    \addlinespace[0.5pt]
    \midrule
    \multicolumn{6}{@{}l@{}}{\textbf{B. Self-certainty}} \\
    \addlinespace[2pt]
AIME 2024 & 0.597 & 0.851 & 0.848 & \sctci{+0.2540}{0.1386}{0.3319} & \cellcolor{rankStable}Yes \\
    \addlinespace[0.5pt]
AIME 2025 & 0.589 & 0.837 & 0.834 & \sctci{+0.2488}{0.1182}{0.3479} & \cellcolor{rankStable}Yes \\
    \addlinespace[0.5pt]
GPQA-Diamond & 0.565 & 0.708 & 0.708 & \sctci{+0.1427}{0.0888}{0.1919} & \cellcolor{rankStable}Yes \\
    \addlinespace[0.5pt]
    \cmidrule(l){1-6}
    \textbf{Overall} & 0.584 & 0.799 & 0.797 & \sctci{+0.2152}{0.1554}{0.2614} & \cellcolor{rankStable}Yes \\
    \addlinespace[0.5pt]
    \midrule
    \multicolumn{6}{@{}l@{}}{\textbf{C. Negative entropy}} \\
    \addlinespace[2pt]
AIME 2024 & 0.491 & 0.751 & 0.749 & \sctci{+0.2605}{0.1661}{0.3337} & \cellcolor{rankStable}Yes \\
    \addlinespace[0.5pt]
AIME 2025 & 0.504 & 0.707 & 0.704 & \sctci{+0.2031}{0.0951}{0.3051} & \cellcolor{rankStable}Yes \\
    \addlinespace[0.5pt]
GPQA-Diamond & 0.534 & 0.668 & 0.668 & \sctci{+0.1342}{0.0799}{0.1836} & \cellcolor{rankStable}Yes \\
    \addlinespace[0.5pt]
    \cmidrule(l){1-6}
    \textbf{Overall} & 0.509 & 0.709 & 0.707 & \sctci{+0.1992}{0.1479}{0.2433} & \cellcolor{rankStable}Yes \\
    \addlinespace[0.5pt]
    \midrule
    \multicolumn{6}{@{}l@{}}{\textbf{D. Maximum log probability}} \\
    \addlinespace[2pt]
AIME 2024 & 0.488 & 0.739 & 0.737 & \sctci{+0.2514}{0.1561}{0.3268} & \cellcolor{rankStable}Yes \\
    \addlinespace[0.5pt]
AIME 2025 & 0.501 & 0.694 & 0.691 & \sctci{+0.1930}{0.0900}{0.2935} & \cellcolor{rankStable}Yes \\
    \addlinespace[0.5pt]
GPQA-Diamond & 0.532 & 0.665 & 0.665 & \sctci{+0.1328}{0.0780}{0.1823} & \cellcolor{rankStable}Yes \\
    \addlinespace[0.5pt]
    \cmidrule(l){1-6}
    \textbf{Overall} & 0.507 & 0.699 & 0.697 & \sctci{+0.1924}{0.1422}{0.2367} & \cellcolor{rankStable}Yes \\
    \addlinespace[0.5pt]
    \bottomrule
  \end{tabular*}
\end{table}

\subsection{Comparing Both Targets on the Same Question Set}
\label{app:common-question-set}
To compare both targets on the same question set, we retain only questions
with both correct and incorrect responses for each model, yielding 865 of
1,290 model--question groups. Within-question AUROC is unchanged;
we recompute cross-question C-Index on this restricted set using the same
evaluation rules.
Tables~\ref{tab:common-question-support-by-model} and
\ref{tab:common-question-support-by-dataset} report the results.

Restricting both targets to the same questions lowers the overall
cross-question C-Index for all four metrics; for logit gap, it falls from
$0.740$ to $0.666$. On AIME 2024 and GPQA-Diamond, Gemma4-26B has lower
cross-question C-Index than within-question AUROC for all four metrics;
difference intervals cross zero. Overall difference intervals include zero for
logit gap and self-certainty but remain above zero for the other two metrics.
Thus, question selection partly accounts for the cross-question advantage
and changes both the empirical-accuracy range and the number of comparable
questions.

\begin{table}[!htbp]
  \centering
  \footnotesize
  \newcommand{\sctci}[3]{\begingroup\renewcommand{\arraystretch}{0.80}%
    \begin{tabular}{@{}c@{}}#1\\[-1.3pt]{[#2,\,#3]}\end{tabular}\endgroup}
  \caption{Discrimination on questions with both correct and incorrect responses, by model. Gemma4-26B reverses the cross-question advantage for all four metrics; difference intervals cross zero.
  Model rows average three datasets equally; Overall averages all 15 settings equally.
  Both targets use questions with both correct and incorrect responses;
  within-question AUROC is unchanged.
  Blue ``Yes'' denotes a positive difference, not statistical significance.
  Brackets give pointwise 95\% paired question-bootstrap intervals
  (Appendix~\ref{app:diagnostic-samples}).}
  \label{tab:common-question-support-by-model}
  \setlength{\tabcolsep}{2.4pt}
  \renewcommand{\arraystretch}{1.08}
  \begin{tabular*}{\linewidth}{@{\extracolsep{\fill}}lcccc@{}}
    \toprule
    \textbf{Model}
    & \shortstack{\textbf{within-question}\\AUROC (W)}
    & \shortstack{\textbf{cross-question}\\C-Index (C)}
    & \shortstack{\textbf{C $-$ W}\\95\% CI}
    & \shortstack{\textbf{cross-question}\\\textbf{$>$ within-question?}} \\
    \midrule
    \multicolumn{5}{@{}l@{}}{\textbf{A. Logit gap ($\Delta_{19}$)}} \\
    \addlinespace[2pt]
\qwenicon\hspace{0.20em}Qwen3-8B & 0.674 & 0.730 & \sctci{+0.0557}{\textminus 0.0106}{0.1220} & \cellcolor{rankStable}Yes \\
    \addlinespace[0.5pt]
\qwenicon\hspace{0.20em}Qwen3-32B & 0.647 & 0.734 & \sctci{+0.0866}{0.0279}{0.1422} & \cellcolor{rankStable}Yes \\
    \addlinespace[0.5pt]
\gemmaicon\hspace{0.20em}Gemma4-4B & 0.656 & 0.745 & \sctci{+0.0896}{0.0045}{0.1729} & \cellcolor{rankStable}Yes \\
    \addlinespace[0.5pt]
\gemmaicon\hspace{0.20em}Gemma4-26B & 0.568 & 0.498 & \sctci{\textminus 0.0697}{\textminus 0.2793}{0.1676} & \cellcolor{white}No \\
    \addlinespace[0.5pt]
\openaiicon\hspace{0.20em}GPT-OSS-20B & 0.529 & 0.621 & \sctci{+0.0921}{0.0165}{0.1648} & \cellcolor{rankStable}Yes \\
    \addlinespace[0.5pt]
    \cmidrule(l){1-5}
    \textbf{Overall} & 0.615 & 0.666 & \sctci{+0.0509}{\textminus 0.0066}{0.1115} & \cellcolor{rankStable}Yes \\
    \addlinespace[0.5pt]
    \midrule
    \multicolumn{5}{@{}l@{}}{\textbf{B. Self-certainty}} \\
    \addlinespace[2pt]
\qwenicon\hspace{0.20em}Qwen3-8B & 0.672 & 0.726 & \sctci{+0.0539}{\textminus 0.0120}{0.1201} & \cellcolor{rankStable}Yes \\
    \addlinespace[0.5pt]
\qwenicon\hspace{0.20em}Qwen3-32B & 0.644 & 0.729 & \sctci{+0.0847}{0.0247}{0.1436} & \cellcolor{rankStable}Yes \\
    \addlinespace[0.5pt]
\gemmaicon\hspace{0.20em}Gemma4-4B & 0.655 & 0.740 & \sctci{+0.0857}{\textminus 0.0012}{0.1700} & \cellcolor{rankStable}Yes \\
    \addlinespace[0.5pt]
\gemmaicon\hspace{0.20em}Gemma4-26B & 0.573 & 0.491 & \sctci{\textminus 0.0820}{\textminus 0.2773}{0.1474} & \cellcolor{white}No \\
    \addlinespace[0.5pt]
\openaiicon\hspace{0.20em}GPT-OSS-20B & 0.529 & 0.619 & \sctci{+0.0898}{0.0162}{0.1629} & \cellcolor{rankStable}Yes \\
    \addlinespace[0.5pt]
    \cmidrule(l){1-5}
    \textbf{Overall} & 0.615 & 0.661 & \sctci{+0.0464}{\textminus 0.0113}{0.1050} & \cellcolor{rankStable}Yes \\
    \addlinespace[0.5pt]
    \midrule
    \multicolumn{5}{@{}l@{}}{\textbf{C. Negative entropy}} \\
    \addlinespace[2pt]
\qwenicon\hspace{0.20em}Qwen3-8B & 0.499 & 0.656 & \sctci{+0.1577}{0.0750}{0.2402} & \cellcolor{rankStable}Yes \\
    \addlinespace[0.5pt]
\qwenicon\hspace{0.20em}Qwen3-32B & 0.502 & 0.639 & \sctci{+0.1369}{0.0643}{0.2034} & \cellcolor{rankStable}Yes \\
    \addlinespace[0.5pt]
\gemmaicon\hspace{0.20em}Gemma4-4B & 0.666 & 0.749 & \sctci{+0.0830}{\textminus 0.0020}{0.1649} & \cellcolor{rankStable}Yes \\
    \addlinespace[0.5pt]
\gemmaicon\hspace{0.20em}Gemma4-26B & 0.482 & 0.405 & \sctci{\textminus 0.0769}{\textminus 0.2332}{0.1222} & \cellcolor{white}No \\
    \addlinespace[0.5pt]
\openaiicon\hspace{0.20em}GPT-OSS-20B & 0.510 & 0.590 & \sctci{+0.0793}{\textminus 0.0023}{0.1611} & \cellcolor{rankStable}Yes \\
    \addlinespace[0.5pt]
    \cmidrule(l){1-5}
    \textbf{Overall} & 0.532 & 0.608 & \sctci{+0.0760}{0.0215}{0.1340} & \cellcolor{rankStable}Yes \\
    \addlinespace[0.5pt]
    \midrule
    \multicolumn{5}{@{}l@{}}{\textbf{D. Maximum log probability}} \\
    \addlinespace[2pt]
\qwenicon\hspace{0.20em}Qwen3-8B & 0.493 & 0.650 & \sctci{+0.1571}{0.0739}{0.2391} & \cellcolor{rankStable}Yes \\
    \addlinespace[0.5pt]
\qwenicon\hspace{0.20em}Qwen3-32B & 0.501 & 0.638 & \sctci{+0.1374}{0.0677}{0.2038} & \cellcolor{rankStable}Yes \\
    \addlinespace[0.5pt]
\gemmaicon\hspace{0.20em}Gemma4-4B & 0.667 & 0.747 & \sctci{+0.0805}{\textminus 0.0057}{0.1630} & \cellcolor{rankStable}Yes \\
    \addlinespace[0.5pt]
\gemmaicon\hspace{0.20em}Gemma4-26B & 0.471 & 0.406 & \sctci{\textminus 0.0651}{\textminus 0.2202}{0.1374} & \cellcolor{white}No \\
    \addlinespace[0.5pt]
\openaiicon\hspace{0.20em}GPT-OSS-20B & 0.508 & 0.586 & \sctci{+0.0785}{\textminus 0.0036}{0.1593} & \cellcolor{rankStable}Yes \\
    \addlinespace[0.5pt]
    \cmidrule(l){1-5}
    \textbf{Overall} & 0.528 & 0.606 & \sctci{+0.0777}{0.0223}{0.1352} & \cellcolor{rankStable}Yes \\
    \addlinespace[0.5pt]
    \bottomrule
  \end{tabular*}
\end{table}

\FloatBarrier
\begin{table}[!htbp]
  \centering
  \footnotesize
  \newcommand{\sctci}[3]{\begingroup\renewcommand{\arraystretch}{0.88}%
    \begin{tabular}{@{}c@{}}#1\\[-1.3pt]{[#2,\,#3]}\end{tabular}\endgroup}
  \caption{Discrimination on questions with both correct and incorrect responses, by dataset. AIME 2024 reverses the cross-question advantage for logit gap and self-certainty; difference intervals cross zero.
  Dataset rows average five models equally; Overall averages all 15 settings equally.
  Both targets use questions with both correct and incorrect responses;
  within-question AUROC is unchanged.
  Blue ``Yes'' denotes a positive difference, not statistical significance.
  Brackets give pointwise 95\% paired question-bootstrap intervals
  (Appendix~\ref{app:diagnostic-samples}).}
  \label{tab:common-question-support-by-dataset}
  \setlength{\tabcolsep}{2.4pt}
  \renewcommand{\arraystretch}{1.08}
  \begin{tabular*}{\linewidth}{@{\extracolsep{\fill}}lcccc@{}}
    \toprule
    \textbf{Dataset}
    & \shortstack{\textbf{within-question}\\AUROC (W)}
    & \shortstack{\textbf{cross-question}\\C-Index (C)}
    & \shortstack{\textbf{C $-$ W}\\95\% CI}
    & \shortstack{\textbf{cross-question}\\\textbf{$>$ within-question?}} \\
    \midrule
    \multicolumn{5}{@{}l@{}}{\textbf{A. Logit gap ($\Delta_{19}$)}} \\
    \addlinespace[2pt]
AIME 2024 & 0.624 & 0.614 & \sctci{\textminus 0.0106}{\textminus 0.1451}{0.1278} & \cellcolor{white}No \\
    \addlinespace[0.5pt]
AIME 2025 & 0.637 & 0.765 & \sctci{+0.1279}{0.0033}{0.2353} & \cellcolor{rankStable}Yes \\
    \addlinespace[0.5pt]
GPQA-Diamond & 0.583 & 0.618 & \sctci{+0.0353}{\textminus 0.0017}{0.0690} & \cellcolor{rankStable}Yes \\
    \addlinespace[0.5pt]
    \cmidrule(l){1-5}
    \textbf{Overall} & 0.615 & 0.666 & \sctci{+0.0509}{\textminus 0.0066}{0.1115} & \cellcolor{rankStable}Yes \\
    \addlinespace[0.5pt]
    \midrule
    \multicolumn{5}{@{}l@{}}{\textbf{B. Self-certainty}} \\
    \addlinespace[2pt]
AIME 2024 & 0.623 & 0.602 & \sctci{\textminus 0.0205}{\textminus 0.1466}{0.1147} & \cellcolor{white}No \\
    \addlinespace[0.5pt]
AIME 2025 & 0.638 & 0.764 & \sctci{+0.1263}{\textminus 0.0018}{0.2339} & \cellcolor{rankStable}Yes \\
    \addlinespace[0.5pt]
GPQA-Diamond & 0.583 & 0.616 & \sctci{+0.0335}{\textminus 0.0039}{0.0675} & \cellcolor{rankStable}Yes \\
    \addlinespace[0.5pt]
    \cmidrule(l){1-5}
    \textbf{Overall} & 0.615 & 0.661 & \sctci{+0.0464}{\textminus 0.0113}{0.1050} & \cellcolor{rankStable}Yes \\
    \addlinespace[0.5pt]
    \midrule
    \multicolumn{5}{@{}l@{}}{\textbf{C. Negative entropy}} \\
    \addlinespace[2pt]
AIME 2024 & 0.522 & 0.561 & \sctci{+0.0385}{\textminus 0.0825}{0.1709} & \cellcolor{rankStable}Yes \\
    \addlinespace[0.5pt]
AIME 2025 & 0.526 & 0.674 & \sctci{+0.1484}{0.0339}{0.2548} & \cellcolor{rankStable}Yes \\
    \addlinespace[0.5pt]
GPQA-Diamond & 0.547 & 0.589 & \sctci{+0.0412}{0.0017}{0.0795} & \cellcolor{rankStable}Yes \\
    \addlinespace[0.5pt]
    \cmidrule(l){1-5}
    \textbf{Overall} & 0.532 & 0.608 & \sctci{+0.0760}{0.0215}{0.1340} & \cellcolor{rankStable}Yes \\
    \addlinespace[0.5pt]
    \midrule
    \multicolumn{5}{@{}l@{}}{\textbf{D. Maximum log probability}} \\
    \addlinespace[2pt]
AIME 2024 & 0.518 & 0.560 & \sctci{+0.0414}{\textminus 0.0812}{0.1750} & \cellcolor{rankStable}Yes \\
    \addlinespace[0.5pt]
AIME 2025 & 0.520 & 0.669 & \sctci{+0.1493}{0.0357}{0.2553} & \cellcolor{rankStable}Yes \\
    \addlinespace[0.5pt]
GPQA-Diamond & 0.546 & 0.588 & \sctci{+0.0423}{0.0024}{0.0802} & \cellcolor{rankStable}Yes \\
    \addlinespace[0.5pt]
    \cmidrule(l){1-5}
    \textbf{Overall} & 0.528 & 0.606 & \sctci{+0.0777}{0.0223}{0.1352} & \cellcolor{rankStable}Yes \\
    \addlinespace[0.5pt]
    \bottomrule
  \end{tabular*}
\end{table}

\subsection{Rank Sensitivity}
\label{app:rank-sensitivity}

We test how competitor rank affects discrimination using $\Delta_k=z_0-z_k$
(Section~\ref{sec:experiment-setup}). Here, $k$ indexes tokens in descending
probability order, so $\Delta_k$ compares the most probable token with the
$(k+1)$th most probable token. This analysis uses a separate response
collection with top-100 logits (Appendix~\ref{app:diagnostic-samples}). At
each evaluated $k$, we average token-level gaps over each complete response
and evaluate both prediction targets on the same responses and labels.
We evaluate 21 values of $k$ between 2 and 99, including $k=19$, which is
used in the main comparison.

\begin{figure}[!t]
  \setlength{\abovecaptionskip}{3pt}
  \centering
  \includegraphics[width=\linewidth]{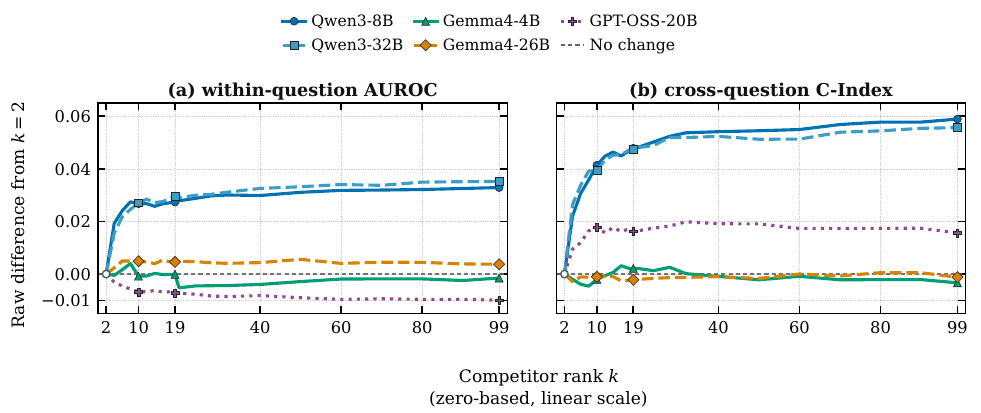}
  \caption{Rank sensitivity varies by model and target. Panels show each model's
  change in within-question AUROC (left) or cross-question C-Index (right)
  relative to $k=2$. Curves show averages across three datasets at each of the 21 evaluated
  values of $k$; markers highlight $k\in\{2,10,19,99\}$.
  Table~\ref{tab:rank-sensitivity} gives absolute scores.}
  \label{fig:competitor-rank-discrimination}
\end{figure}

From $k=2$ to
$k=99$, within-question AUROC and cross-question C-Index both increase
for the Qwen models and change little for the Gemma models. For
GPT-OSS-20B, within-question AUROC decreases, whereas cross-question
C-Index increases. Across settings, increasing $k$
from 2 to 10 raises mean within-question AUROC from $0.614$ to $0.625$ and
mean cross-question C-Index from $0.738$ to $0.757$. Both averages change little beyond this competitor rank.

\begin{table}[!htbp]
  \centering
  \footnotesize
  \caption{Absolute logit-gap discrimination at selected competitor ranks.
  Cells report mean within-question AUROC (W) and cross-question C-Index (C)
  over three datasets per model; Overall averages all 15 settings.
  Light blue identifies $k=19$, corresponding to the 20th most probable token
  used in the main comparison.
  Bold marks the highest W and C separately within each column across
  displayed indices, including ties at three decimal places. The response collection differs from that used in the main four-metric
  comparison in Figure~\ref{fig:target-comparison}.}
  \label{tab:rank-sensitivity}
  \setlength{\tabcolsep}{2.4pt}
  \renewcommand{\arraystretch}{1.08}
  \begin{tabular}{@{}r@{\hspace{7pt}}ccccc@{\hspace{8pt}}c@{}}
    \toprule
    \multirow{2}{*}{\textbf{Index $k$}}
    & \multicolumn{5}{c}{\textbf{Per-model average (W/C)}}
    & \multicolumn{1}{c}{\textbf{Overall}} \\
    \cmidrule(lr){2-6}\cmidrule(l){7-7}
    & \qwenicon\hspace{0.20em}Qwen3-8B & \qwenicon\hspace{0.20em}Qwen3-32B
    & \gemmaicon\hspace{0.20em}Gemma4-4B & \gemmaicon\hspace{0.20em}Gemma4-26B
    & \openaiicon\hspace{0.20em}GPT-OSS-20B & W / C \\
    \midrule
    2 & 0.635 / 0.745 & 0.642 / 0.721 & 0.699 / 0.751 & 0.590 / \textbf{0.790} & \textbf{0.506} / 0.685 & 0.614 / 0.738 \\
    6 & 0.659 / 0.775 & 0.664 / 0.755 & \textbf{0.701} / 0.747 & \textbf{0.595} / 0.789 & 0.502 / 0.697 & 0.624 / 0.753 \\
    10 & 0.661 / 0.786 & 0.669 / 0.760 & 0.698 / 0.749 & \textbf{0.595} / 0.789 & 0.500 / 0.703 & 0.625 / 0.757 \\
    14 & 0.660 / 0.791 & 0.669 / 0.766 & 0.699 / 0.752 & 0.594 / 0.789 & 0.500 / 0.703 & 0.625 / 0.760 \\
    18 & 0.662 / 0.791 & 0.671 / 0.768 & 0.699 / \textbf{0.753} & 0.594 / 0.788 & 0.500 / 0.702 & 0.625 / 0.761 \\
    \rowcolor{rankStable}
    19 & 0.662 / 0.792 & 0.672 / 0.768 & 0.699 / \textbf{0.753} & 0.594 / 0.788 & 0.499 / 0.702 & 0.625 / 0.761 \\
    20 & 0.663 / 0.793 & 0.671 / 0.769 & 0.694 / \textbf{0.753} & \textbf{0.595} / 0.788 & 0.499 / 0.702 & 0.624 / 0.761 \\
    40 & 0.665 / 0.799 & 0.675 / 0.773 & 0.695 / 0.750 & 0.594 / 0.789 & 0.498 / \textbf{0.704} & 0.625 / 0.763 \\
    60 & 0.666 / 0.799 & 0.676 / 0.772 & 0.697 / 0.750 & 0.594 / \textbf{0.790} & 0.497 / 0.703 & \textbf{0.626} / 0.763 \\
    80 & 0.667 / 0.802 & \textbf{0.677} / 0.775 & 0.697 / 0.749 & 0.594 / \textbf{0.790} & 0.497 / 0.703 & \textbf{0.626} / \textbf{0.764} \\
    99 & \textbf{0.668} / \textbf{0.803} & \textbf{0.677} / \textbf{0.777} & 0.697 / 0.748 & 0.594 / 0.789 & 0.496 / 0.701 & \textbf{0.626} / \textbf{0.764} \\
    \bottomrule
  \end{tabular}
\end{table}

\FloatBarrier
\Needspace*{7\baselineskip}
\section{Details for Section 4.2}
\label{app:certainty-noise-details}

\subsection{Token Categories and Positions Within Words}
\label{app:surface-form-diagnostics}

We compute logit gap $\Delta_{19}$, self-certainty, negative entropy, and
maximum log probability using the definitions in
Section~\ref{sec:experiment-setup} and Equation~\ref{eq:certainty-metrics},
with $K=20$.
The local diagnostics use all sampled responses from the 15 model--dataset
settings.

We assign tokens to five categories using their decoded characters.
Structural comprises empty strings and tokens containing only whitespace,
markup, or punctuation. Among the remaining strings, those containing digits
are labeled Digit. Mathematical or operator symbols are labeled Symbolic. Other alphabetic strings
are Semantic; residual cases are Others. These character-based categories
do not specify a token's role in reasoning.

We group tokens into alphabetic words using heuristic boundaries based on
the decoded characters. Each token within a word is a subword, or
piece. We analyze two-, three-, and four-piece words separately, with each
word occurrence contributing one observation at every position.
Table~\ref{tab:local-group-coverage} reports the observed coverage.

\begin{table}[htbp]
\centering
\caption{Coverage by token category and word length for all sampled responses (five models, three datasets). Occurrences count tokens by category and words by position group. Symbolic excludes digits; Digit contains digits. Traces and model--question pairs can belong to multiple groups; positions in a word share coverage.}
\label{tab:local-group-coverage}
\small
\setlength{\tabcolsep}{3pt}
\renewcommand{\arraystretch}{1.04}
\begin{tabular}{@{}lrrrr@{}}
\toprule
Group & Occurrences & Traces & Model--questions & Settings \\
\midrule
Semantic & 2,758,980,303 & 660,480 & 1,290 & 15/15 \\
Structural & 1,535,677,651 & 660,480 & 1,290 & 15/15 \\
Symbolic & 1,311,805,321 & 660,480 & 1,290 & 15/15 \\
Digit & 994,217,932 & 652,510 & 1,288 & 15/15 \\
Others & 25,618,508 & 328,951 & 873 & 11/15 \\
2-piece words & 136,824,127 & 655,934 & 1,290 & 15/15 \\
3-piece words & 34,707,433 & 554,108 & 1,279 & 15/15 \\
4-piece words & 11,382,863 & 340,353 & 1,075 & 15/15 \\
\bottomrule
\end{tabular}
\end{table}

Within each setting, we average token scores by category and compare each
category mean with the Semantic mean. For all four metrics, Structural,
Symbolic, and Digit tokens have higher mean certainty than Semantic tokens
in every setting. For each metric, the 95\% paired question-bootstrap intervals for all
45 category contrasts are above zero.
Table~\ref{tab:token-category-by-model} reports the raw means.

Table~\ref{tab:type-equal-by-model} compares the all-token mean of
$\Delta_{19}$ (Original) with an equal-weight average of the Semantic,
Structural, Symbolic, and Digit means (Equal), excluding Others from Equal.
For cross-question scores, we pool tokens across responses to each question.
Both use matched samples under the eligibility rules in
Appendix~\ref{app:diagnostic-samples}. For most models, Equal improves
discrimination on both targets.

\begin{table}[!htbp]
  \centering
  \begin{minipage}{0.8\linewidth}
  \footnotesize
  \captionsetup{font=footnotesize,skip=3pt}
  \caption{Equal category weighting improves discrimination on both targets for most models. Original averages all token scores; Equal assigns 25\% to each of the Semantic, Structural, Symbolic, and Digit means, excluding Others. Both use logit gap $\Delta_{19}$ with the scoring and sample-selection rules in Appendix~\ref{app:surface-form-diagnostics}. Entries are equally weighted averages across the three datasets. Bold marks the higher value, including ties.}
  \label{tab:type-equal-by-model}
  \setlength{\tabcolsep}{2pt}
  \renewcommand{\arraystretch}{1.10}
  \begin{tabular*}{\linewidth}{@{\extracolsep{\fill}}l r r r r@{}}
    \toprule
    \multirow{2}{*}{\textbf{Model}}
    & \multicolumn{2}{c}{\shortstack{\textbf{cross-question}\\C-Index $\uparrow$}}
    & \multicolumn{2}{c}{\shortstack{\textbf{within-question}\\AUROC $\uparrow$}} \\
    \cmidrule(lr){2-3}\cmidrule(l){4-5}
    & Original & Equal & Original & Equal \\
    \midrule
    \qwenicon\hspace{0.20em}Qwen3-8B
      & 0.7673 & \textbf{0.8070} & 0.6739 & \textbf{0.6842} \\
    \qwenicon\hspace{0.20em}Qwen3-32B
      & 0.7720 & \textbf{0.7926} & 0.6467 & \textbf{0.6552} \\
    \addlinespace[3pt]
    \gemmaicon\hspace{0.20em}Gemma 4-E4B
      & 0.7717 & \textbf{0.7962} & \textbf{0.6556} & 0.6481 \\
    \gemmaicon\hspace{0.20em}Gemma4-26B
      & 0.7236 & \textbf{0.7240} & 0.5675 & \textbf{0.5907} \\
    \addlinespace[3pt]
    \openaiicon\hspace{0.20em}GPT-OSS-20B
      & 0.6546 & \textbf{0.6819} & 0.5307 & \textbf{0.5498} \\
    \bottomrule
  \end{tabular*}
  \end{minipage}
\end{table}

For word positions, we average scores by word length and position within
each setting and compare each later position with the first.
Later pieces have higher mean certainty in all 90 comparisons per metric:
six later positions across the three word lengths in 15 settings.
All paired intervals are above zero for negative entropy and maximum log
probability. Logit gap and self-certainty each have two intervals crossing
zero, all involving Gemma4-26B. Certainty need not increase monotonically
with subword position. These associations are consistent with constraints on word
completion, although tokenization and word context are not separately
controlled. Table~\ref{tab:subword-position-by-model} reports the position
means; Appendix~\ref{app:diagnostic-samples} describes how the intervals are computed.

\begin{table}[!htbp]
  \centering
  \footnotesize
  \renewcommand{\localmean}[2]{\cellcolor{localCertaintyShade!#1!white}#2}
  \caption{Token-category means for four certainty metrics. For each model shown, Structural, Symbolic, and Digit means exceed the Semantic mean. Means use all sampled responses, with equal weight for token occurrences within each setting. Model rows average three datasets equally; Overall averages 15 settings. Symbolic excludes digits; Digit contains digits. NA denotes a missing category in at least one required setting. Darker shading denotes larger raw means; each metric shares a color scale with Table~\ref{tab:subword-position-by-model}.}
  \label{tab:token-category-by-model}
  \setlength{\tabcolsep}{2.4pt}
  \renewcommand{\arraystretch}{1.08}
  \begin{tabular}{@{}>{\raggedright\arraybackslash}p{\dimexpr.25\linewidth-2.5\tabcolsep\relax}*{5}{>{\raggedleft\arraybackslash}p{\dimexpr.15\linewidth-1.5\tabcolsep\relax}}@{}}
    \toprule
    \multirow{2}{*}{\textbf{Model}} & \multicolumn{5}{c}{\textbf{Token category}} \\
    \cmidrule(l){2-6}
    & Semantic & Structural & Symbolic & Digit & Others \\
    \midrule
    \multicolumn{6}{@{}l@{}}{\makebox[\linewidth][l]{\textbf{A. Logit gap}~($\Delta_{19}$)\hfill Color scale: 8.159~\colorbox{localCertaintyShade!0!white}{\strut\hspace{5pt}}\colorbox{localCertaintyShade!10.5!white}{\strut\hspace{5pt}}\colorbox{localCertaintyShade!21!white}{\strut\hspace{5pt}}\colorbox{localCertaintyShade!31.5!white}{\strut\hspace{5pt}}\colorbox{localCertaintyShade!42!white}{\strut\hspace{5pt}}~29.115}} \\
    \addlinespace[3pt]
    \qwenicon\hspace{0.20em}Qwen3-8B & \localmean{17.31}{16.794} & \localmean{26.91}{21.585} & \localmean{28.62}{22.441} & \localmean{39.37}{27.802} & \localmean{35.68}{25.962} \\
    \qwenicon\hspace{0.20em}Qwen3-32B & \localmean{18.60}{17.439} & \localmean{28.10}{22.178} & \localmean{29.29}{22.772} & \localmean{37.49}{26.863} & \localmean{32.95}{24.600} \\
    \gemmaicon\hspace{0.20em}Gemma 4-E4B & \localmean{9.59}{12.946} & \localmean{18.15}{17.217} & \localmean{21.28}{18.775} & \localmean{27.10}{21.679} & NA \\
    \gemmaicon\hspace{0.20em}Gemma4-26B & \localmean{16.01}{16.147} & \localmean{21.35}{18.811} & \localmean{22.96}{19.616} & \localmean{26.74}{21.501} & NA \\
    \openaiicon\hspace{0.20em}GPT-OSS-20B & \localmean{3.18}{9.745} & \localmean{8.99}{12.647} & \localmean{12.89}{14.591} & \localmean{18.66}{17.469} & \localmean{18.07}{17.174} \\
    \cmidrule(l){1-6}
    \textbf{Overall} & \localmean{12.94}{14.614} & \localmean{20.70}{18.488} & \localmean{23.01}{19.639} & \localmean{29.87}{23.063} & NA \\
    \midrule
    \multicolumn{6}{@{}l@{}}{\makebox[\linewidth][l]{\textbf{B. Self-certainty}\hfill Color scale: 6.308~\colorbox{localCertaintyShade!0!white}{\strut\hspace{5pt}}\colorbox{localCertaintyShade!10.5!white}{\strut\hspace{5pt}}\colorbox{localCertaintyShade!21!white}{\strut\hspace{5pt}}\colorbox{localCertaintyShade!31.5!white}{\strut\hspace{5pt}}\colorbox{localCertaintyShade!42!white}{\strut\hspace{5pt}}~24.253}} \\
    \addlinespace[3pt]
    \qwenicon\hspace{0.20em}Qwen3-8B & \localmean{15.72}{13.025} & \localmean{26.71}{17.720} & \localmean{28.94}{18.672} & \localmean{39.08}{23.007} & \localmean{35.82}{21.614} \\
    \qwenicon\hspace{0.20em}Qwen3-32B & \localmean{16.64}{13.417} & \localmean{27.66}{18.124} & \localmean{29.38}{18.863} & \localmean{38.45}{22.737} & \localmean{32.98}{20.398} \\
    \gemmaicon\hspace{0.20em}Gemma 4-E4B & \localmean{9.96}{10.565} & \localmean{18.97}{14.415} & \localmean{22.23}{15.807} & \localmean{28.55}{18.507} & NA \\
    \gemmaicon\hspace{0.20em}Gemma4-26B & \localmean{17.21}{13.660} & \localmean{22.65}{15.987} & \localmean{23.96}{16.545} & \localmean{28.36}{18.424} & NA \\
    \openaiicon\hspace{0.20em}GPT-OSS-20B & \localmean{3.41}{7.764} & \localmean{9.32}{10.292} & \localmean{13.23}{11.962} & \localmean{19.57}{14.669} & \localmean{18.19}{14.082} \\
    \cmidrule(l){1-6}
    \textbf{Overall} & \localmean{12.59}{11.686} & \localmean{21.06}{15.308} & \localmean{23.55}{16.370} & \localmean{30.80}{19.469} & NA \\
    \midrule
    \multicolumn{6}{@{}l@{}}{\makebox[\linewidth][l]{\textbf{C. Negative entropy}\hfill Color scale: \textminus 1.2735~\colorbox{localCertaintyShade!0!white}{\strut\hspace{5pt}}\colorbox{localCertaintyShade!10.5!white}{\strut\hspace{5pt}}\colorbox{localCertaintyShade!21!white}{\strut\hspace{5pt}}\colorbox{localCertaintyShade!31.5!white}{\strut\hspace{5pt}}\colorbox{localCertaintyShade!42!white}{\strut\hspace{5pt}}~\textminus 0.0029}} \\
    \addlinespace[3pt]
    \qwenicon\hspace{0.20em}Qwen3-8B & \localmean{26.75}{\textminus 0.4643} & \localmean{34.89}{\textminus 0.2181} & \localmean{37.25}{\textminus 0.1465} & \localmean{41.12}{\textminus 0.0295} & \localmean{39.17}{\textminus 0.0884} \\
    \qwenicon\hspace{0.20em}Qwen3-32B & \localmean{26.34}{\textminus 0.4767} & \localmean{34.44}{\textminus 0.2315} & \localmean{36.85}{\textminus 0.1585} & \localmean{41.17}{\textminus 0.0278} & \localmean{39.08}{\textminus 0.0911} \\
    \gemmaicon\hspace{0.20em}Gemma 4-E4B & \localmean{23.17}{\textminus 0.5726} & \localmean{35.27}{\textminus 0.2066} & \localmean{37.90}{\textminus 0.1269} & \localmean{40.73}{\textminus 0.0412} & NA \\
    \gemmaicon\hspace{0.20em}Gemma4-26B & \localmean{33.10}{\textminus 0.2721} & \localmean{38.58}{\textminus 0.1062} & \localmean{39.36}{\textminus 0.0827} & \localmean{41.23}{\textminus 0.0261} & NA \\
    \openaiicon\hspace{0.20em}GPT-OSS-20B & \localmean{12.70}{\textminus 0.8892} & \localmean{28.02}{\textminus 0.4257} & \localmean{32.14}{\textminus 0.3013} & \localmean{38.49}{\textminus 0.1090} & \localmean{32.80}{\textminus 0.2812} \\
    \cmidrule(l){1-6}
    \textbf{Overall} & \localmean{24.41}{\textminus 0.5350} & \localmean{34.24}{\textminus 0.2376} & \localmean{36.70}{\textminus 0.1632} & \localmean{40.55}{\textminus 0.0467} & NA \\
    \midrule
    \multicolumn{6}{@{}l@{}}{\makebox[\linewidth][l]{\textbf{D. Maximum log probability}\hfill Color scale: \textminus 0.6874~\colorbox{localCertaintyShade!0!white}{\strut\hspace{5pt}}\colorbox{localCertaintyShade!10.5!white}{\strut\hspace{5pt}}\colorbox{localCertaintyShade!21!white}{\strut\hspace{5pt}}\colorbox{localCertaintyShade!31.5!white}{\strut\hspace{5pt}}\colorbox{localCertaintyShade!42!white}{\strut\hspace{5pt}}~\textminus 0.0011}} \\
    \addlinespace[3pt]
    \qwenicon\hspace{0.20em}Qwen3-8B & \localmean{27.99}{\textminus 0.2300} & \localmean{35.48}{\textminus 0.1076} & \localmean{37.67}{\textminus 0.0718} & \localmean{41.19}{\textminus 0.0143} & \localmean{39.43}{\textminus 0.0431} \\
    \qwenicon\hspace{0.20em}Qwen3-32B & \localmean{27.41}{\textminus 0.2396} & \localmean{34.94}{\textminus 0.1165} & \localmean{37.25}{\textminus 0.0788} & \localmean{41.23}{\textminus 0.0136} & \localmean{39.34}{\textminus 0.0446} \\
    \gemmaicon\hspace{0.20em}Gemma 4-E4B & \localmean{24.87}{\textminus 0.2811} & \localmean{36.33}{\textminus 0.0938} & \localmean{38.44}{\textminus 0.0593} & \localmean{40.90}{\textminus 0.0191} & NA \\
    \gemmaicon\hspace{0.20em}Gemma4-26B & \localmean{33.90}{\textminus 0.1335} & \localmean{39.06}{\textminus 0.0492} & \localmean{39.67}{\textminus 0.0391} & \localmean{41.32}{\textminus 0.0122} & NA \\
    \openaiicon\hspace{0.20em}GPT-OSS-20B & \localmean{15.50}{\textminus 0.4341} & \localmean{30.24}{\textminus 0.1933} & \localmean{33.60}{\textminus 0.1383} & \localmean{38.97}{\textminus 0.0506} & \localmean{33.78}{\textminus 0.1354} \\
    \cmidrule(l){1-6}
    \textbf{Overall} & \localmean{25.93}{\textminus 0.2636} & \localmean{35.21}{\textminus 0.1121} & \localmean{37.33}{\textminus 0.0775} & \localmean{40.72}{\textminus 0.0220} & NA \\
    \bottomrule
  \end{tabular}
\end{table}

\FloatBarrier
\begin{table}[!htbp]
  \centering
  \footnotesize
  \renewcommand{\localmean}[2]{\cellcolor{localCertaintyShade!#1!white}#2}
  \caption{Later subwords have higher mean certainty than the first across metrics and models, although certainty need not increase monotonically with position. Cells report raw means over word occurrences in all sampled responses, separately for two-, three-, and four-piece words. Pos. denotes the position of a subword within a word. Model rows average three datasets equally; Overall averages 15 settings. Darker shading denotes larger raw means; each metric shares a color scale with Table~\ref{tab:token-category-by-model}.}
  \label{tab:subword-position-by-model}
  \setlength{\tabcolsep}{2.4pt}
  \renewcommand{\arraystretch}{1.08}
  \begin{tabular}{@{}>{\raggedright\arraybackslash}p{.18\linewidth}@{\hspace{7pt}}*{9}{>{\raggedleft\arraybackslash}p{\dimexpr(\linewidth-.18\linewidth-16\tabcolsep-7pt)/9\relax}}@{}}
    \toprule
    \multirow{2}{*}{\textbf{Model}} & \multicolumn{2}{c}{\textbf{2-piece words}} & \multicolumn{3}{c}{\textbf{3-piece words}} & \multicolumn{4}{c}{\textbf{4-piece words}} \\
    \cmidrule(lr){2-3}\cmidrule(lr){4-6}\cmidrule(l){7-10}
    & Pos.~1 & Pos.~2 & Pos.~1 & Pos.~2 & Pos.~3 & Pos.~1 & Pos.~2 & Pos.~3 & Pos.~4 \\
    \midrule
    \multicolumn{10}{@{}l@{}}{\makebox[\linewidth][l]{\textbf{A. Logit gap}~($\Delta_{19}$)\hfill Color scale: 8.159~\colorbox{localCertaintyShade!0!white}{\strut\hspace{5pt}}\colorbox{localCertaintyShade!10.5!white}{\strut\hspace{5pt}}\colorbox{localCertaintyShade!21!white}{\strut\hspace{5pt}}\colorbox{localCertaintyShade!31.5!white}{\strut\hspace{5pt}}\colorbox{localCertaintyShade!42!white}{\strut\hspace{5pt}}~29.115}} \\
    \addlinespace[3pt]
    \qwenicon\hspace{0.20em}Qwen3-8B & \localmean{18.18}{17.231} & \localmean{34.34}{25.294} & \localmean{16.97}{16.626} & \localmean{31.96}{24.103} & \localmean{35.94}{26.091} & \localmean{18.19}{17.233} & \localmean{35.04}{25.640} & \localmean{35.44}{25.841} & \localmean{36.57}{26.407} \\
    \qwenicon\hspace{0.20em}Qwen3-32B & \localmean{18.66}{17.471} & \localmean{30.95}{23.601} & \localmean{16.32}{16.301} & \localmean{27.76}{22.008} & \localmean{30.74}{23.497} & \localmean{17.47}{16.878} & \localmean{28.70}{22.479} & \localmean{28.24}{22.248} & \localmean{29.72}{22.988} \\
    \gemmaicon\hspace{0.20em}Gemma 4-E4B & \localmean{8.19}{12.247} & \localmean{21.66}{18.966} & \localmean{5.98}{11.141} & \localmean{20.30}{18.288} & \localmean{24.41}{20.339} & \localmean{5.47}{10.886} & \localmean{19.10}{17.691} & \localmean{22.48}{19.377} & \localmean{23.11}{19.690} \\
    \gemmaicon\hspace{0.20em}Gemma4-26B & \localmean{18.52}{17.397} & \localmean{26.29}{21.277} & \localmean{18.74}{17.512} & \localmean{24.97}{20.619} & \localmean{27.95}{22.103} & \localmean{17.95}{17.116} & \localmean{23.03}{19.651} & \localmean{25.73}{20.999} & \localmean{26.55}{21.404} \\
    \openaiicon\hspace{0.20em}GPT-OSS-20B & \localmean{5.78}{11.042} & \localmean{19.16}{17.717} & \localmean{2.42}{9.367} & \localmean{17.52}{16.900} & \localmean{20.85}{18.560} & \localmean{4.08}{10.194} & \localmean{16.87}{16.576} & \localmean{19.63}{17.952} & \localmean{21.59}{18.934} \\
    \cmidrule(l){1-10}
    \textbf{Overall} & \localmean{13.87}{15.078} & \localmean{26.48}{21.371} & \localmean{12.09}{14.189} & \localmean{24.50}{20.384} & \localmean{27.98}{22.118} & \localmean{12.63}{14.461} & \localmean{24.55}{20.407} & \localmean{26.30}{21.284} & \localmean{27.51}{21.885} \\
    \midrule
    \multicolumn{10}{@{}l@{}}{\makebox[\linewidth][l]{\textbf{B. Self-certainty}\hfill Color scale: 6.308~\colorbox{localCertaintyShade!0!white}{\strut\hspace{5pt}}\colorbox{localCertaintyShade!10.5!white}{\strut\hspace{5pt}}\colorbox{localCertaintyShade!21!white}{\strut\hspace{5pt}}\colorbox{localCertaintyShade!31.5!white}{\strut\hspace{5pt}}\colorbox{localCertaintyShade!42!white}{\strut\hspace{5pt}}~24.253}} \\
    \addlinespace[3pt]
    \qwenicon\hspace{0.20em}Qwen3-8B & \localmean{17.27}{13.688} & \localmean{35.48}{21.466} & \localmean{15.75}{13.038} & \localmean{33.07}{20.440} & \localmean{36.70}{21.990} & \localmean{16.97}{13.559} & \localmean{36.43}{21.872} & \localmean{36.95}{22.097} & \localmean{37.51}{22.335} \\
    \qwenicon\hspace{0.20em}Qwen3-32B & \localmean{17.36}{13.726} & \localmean{32.11}{20.026} & \localmean{14.71}{12.595} & \localmean{28.94}{18.671} & \localmean{31.70}{19.850} & \localmean{15.99}{13.139} & \localmean{30.03}{19.139} & \localmean{29.70}{18.999} & \localmean{30.67}{19.414} \\
    \gemmaicon\hspace{0.20em}Gemma 4-E4B & \localmean{8.49}{9.936} & \localmean{22.42}{15.887} & \localmean{6.30}{8.999} & \localmean{20.57}{15.098} & \localmean{25.18}{17.065} & \localmean{5.52}{8.666} & \localmean{19.07}{14.456} & \localmean{22.30}{15.834} & \localmean{23.40}{16.307} \\
    \gemmaicon\hspace{0.20em}Gemma4-26B & \localmean{19.66}{14.707} & \localmean{28.37}{18.428} & \localmean{19.93}{14.825} & \localmean{26.96}{17.827} & \localmean{30.01}{19.130} & \localmean{18.16}{14.066} & \localmean{24.35}{16.712} & \localmean{27.17}{17.917} & \localmean{28.57}{18.513} \\
    \openaiicon\hspace{0.20em}GPT-OSS-20B & \localmean{6.07}{8.903} & \localmean{20.28}{14.973} & \localmean{2.69}{7.459} & \localmean{18.49}{14.209} & \localmean{21.95}{15.689} & \localmean{4.28}{8.135} & \localmean{18.19}{14.081} & \localmean{21.25}{15.386} & \localmean{22.91}{16.096} \\
    \cmidrule(l){1-10}
    \textbf{Overall} & \localmean{13.77}{12.192} & \localmean{27.73}{18.156} & \localmean{11.88}{11.383} & \localmean{25.61}{17.249} & \localmean{29.11}{18.745} & \localmean{12.18}{11.513} & \localmean{25.61}{17.252} & \localmean{27.47}{18.046} & \localmean{28.61}{18.533} \\
    \midrule
    \multicolumn{10}{@{}l@{}}{\makebox[\linewidth][l]{\textbf{C. Negative entropy}\hfill Color scale: \textminus 1.2735~\colorbox{localCertaintyShade!0!white}{\strut\hspace{5pt}}\colorbox{localCertaintyShade!10.5!white}{\strut\hspace{5pt}}\colorbox{localCertaintyShade!21!white}{\strut\hspace{5pt}}\colorbox{localCertaintyShade!31.5!white}{\strut\hspace{5pt}}\colorbox{localCertaintyShade!42!white}{\strut\hspace{5pt}}~\textminus 0.0029}} \\
    \addlinespace[3pt]
    \qwenicon\hspace{0.20em}Qwen3-8B & \localmean{30.40}{\textminus 0.3538} & \localmean{41.05}{\textminus 0.0316} & \localmean{28.58}{\textminus 0.4088} & \localmean{41.14}{\textminus 0.0290} & \localmean{41.12}{\textminus 0.0295} & \localmean{29.35}{\textminus 0.3857} & \localmean{41.80}{\textminus 0.0089} & \localmean{41.54}{\textminus 0.0167} & \localmean{41.48}{\textminus 0.0187} \\
    \qwenicon\hspace{0.20em}Qwen3-32B & \localmean{29.87}{\textminus 0.3700} & \localmean{40.92}{\textminus 0.0355} & \localmean{27.21}{\textminus 0.4502} & \localmean{41.07}{\textminus 0.0310} & \localmean{41.03}{\textminus 0.0323} & \localmean{28.49}{\textminus 0.4117} & \localmean{41.71}{\textminus 0.0116} & \localmean{41.54}{\textminus 0.0167} & \localmean{41.33}{\textminus 0.0232} \\
    \gemmaicon\hspace{0.20em}Gemma 4-E4B & \localmean{21.78}{\textminus 0.6146} & \localmean{39.47}{\textminus 0.0795} & \localmean{17.00}{\textminus 0.7593} & \localmean{36.01}{\textminus 0.1840} & \localmean{40.98}{\textminus 0.0337} & \localmean{12.42}{\textminus 0.8977} & \localmean{38.26}{\textminus 0.1160} & \localmean{39.50}{\textminus 0.0786} & \localmean{37.10}{\textminus 0.1512} \\
    \gemmaicon\hspace{0.20em}Gemma4-26B & \localmean{34.24}{\textminus 0.2376} & \localmean{41.27}{\textminus 0.0249} & \localmean{34.86}{\textminus 0.2188} & \localmean{41.21}{\textminus 0.0267} & \localmean{41.81}{\textminus 0.0085} & \localmean{31.69}{\textminus 0.3147} & \localmean{38.94}{\textminus 0.0954} & \localmean{40.29}{\textminus 0.0545} & \localmean{40.62}{\textminus 0.0445} \\
    \openaiicon\hspace{0.20em}GPT-OSS-20B & \localmean{20.26}{\textminus 0.6606} & \localmean{39.45}{\textminus 0.0800} & \localmean{12.80}{\textminus 0.8863} & \localmean{37.37}{\textminus 0.1430} & \localmean{39.97}{\textminus 0.0641} & \localmean{19.06}{\textminus 0.6970} & \localmean{40.03}{\textminus 0.0626} & \localmean{40.37}{\textminus 0.0521} & \localmean{40.37}{\textminus 0.0523} \\
    \cmidrule(l){1-10}
    \textbf{Overall} & \localmean{27.31}{\textminus 0.4473} & \localmean{40.43}{\textminus 0.0503} & \localmean{24.09}{\textminus 0.5447} & \localmean{39.36}{\textminus 0.0827} & \localmean{40.98}{\textminus 0.0336} & \localmean{24.20}{\textminus 0.5414} & \localmean{40.15}{\textminus 0.0589} & \localmean{40.65}{\textminus 0.0437} & \localmean{40.18}{\textminus 0.0580} \\
    \midrule
    \multicolumn{10}{@{}l@{}}{\makebox[\linewidth][l]{\textbf{D. Maximum log probability}\hfill Color scale: \textminus 0.6874~\colorbox{localCertaintyShade!0!white}{\strut\hspace{5pt}}\colorbox{localCertaintyShade!10.5!white}{\strut\hspace{5pt}}\colorbox{localCertaintyShade!21!white}{\strut\hspace{5pt}}\colorbox{localCertaintyShade!31.5!white}{\strut\hspace{5pt}}\colorbox{localCertaintyShade!42!white}{\strut\hspace{5pt}}~\textminus 0.0011}} \\
    \addlinespace[3pt]
    \qwenicon\hspace{0.20em}Qwen3-8B & \localmean{31.57}{\textminus 0.1715} & \localmean{41.15}{\textminus 0.0149} & \localmean{29.95}{\textminus 0.1979} & \localmean{41.18}{\textminus 0.0145} & \localmean{41.19}{\textminus 0.0143} & \localmean{30.43}{\textminus 0.1902} & \localmean{41.82}{\textminus 0.0040} & \localmean{41.58}{\textminus 0.0079} & \localmean{41.54}{\textminus 0.0087} \\
    \qwenicon\hspace{0.20em}Qwen3-32B & \localmean{30.93}{\textminus 0.1819} & \localmean{41.03}{\textminus 0.0170} & \localmean{28.60}{\textminus 0.2201} & \localmean{41.13}{\textminus 0.0153} & \localmean{41.12}{\textminus 0.0155} & \localmean{29.71}{\textminus 0.2019} & \localmean{41.74}{\textminus 0.0053} & \localmean{41.58}{\textminus 0.0080} & \localmean{41.39}{\textminus 0.0111} \\
    \gemmaicon\hspace{0.20em}Gemma 4-E4B & \localmean{23.17}{\textminus 0.3088} & \localmean{39.86}{\textminus 0.0360} & \localmean{19.29}{\textminus 0.3722} & \localmean{36.47}{\textminus 0.0914} & \localmean{41.16}{\textminus 0.0148} & \localmean{13.12}{\textminus 0.4731} & \localmean{39.07}{\textminus 0.0490} & \localmean{39.90}{\textminus 0.0354} & \localmean{37.61}{\textminus 0.0729} \\
    \gemmaicon\hspace{0.20em}Gemma4-26B & \localmean{34.98}{\textminus 0.1159} & \localmean{41.34}{\textminus 0.0119} & \localmean{35.58}{\textminus 0.1060} & \localmean{41.17}{\textminus 0.0146} & \localmean{41.84}{\textminus 0.0038} & \localmean{32.54}{\textminus 0.1556} & \localmean{39.61}{\textminus 0.0402} & \localmean{40.74}{\textminus 0.0218} & \localmean{40.62}{\textminus 0.0236} \\
    \openaiicon\hspace{0.20em}GPT-OSS-20B & \localmean{22.58}{\textminus 0.3184} & \localmean{39.97}{\textminus 0.0342} & \localmean{15.49}{\textminus 0.4342} & \localmean{37.97}{\textminus 0.0670} & \localmean{40.41}{\textminus 0.0271} & \localmean{20.98}{\textminus 0.3446} & \localmean{40.41}{\textminus 0.0271} & \localmean{40.73}{\textminus 0.0218} & \localmean{40.64}{\textminus 0.0233} \\
    \cmidrule(l){1-10}
    \textbf{Overall} & \localmean{28.65}{\textminus 0.2193} & \localmean{40.67}{\textminus 0.0228} & \localmean{25.78}{\textminus 0.2661} & \localmean{39.58}{\textminus 0.0406} & \localmean{41.14}{\textminus 0.0151} & \localmean{25.36}{\textminus 0.2731} & \localmean{40.53}{\textminus 0.0251} & \localmean{40.91}{\textminus 0.0190} & \localmean{40.36}{\textminus 0.0279} \\
    \bottomrule
  \end{tabular}
\end{table}

\subsection{Can Simple Token Masks Improve Correctness Ranking?}
\label{app:mask-noisy-token}

We keep the generated traces fixed and compare token masks that select
the positions used to compute mean certainty. All-token (All in the tables)
uses every position. First-piece retains only the first piece of each detected
multi-token word or number and leaves other tokens unchanged. Exclude Digit
and Others removes tokens assigned to either category. We average the
retained scores for each condition.

We use the same eligible responses for all metrics and masks
(Appendix~\ref{app:diagnostic-samples}). We evaluate within-question AUROC
and cross-question C-Index as defined in Section~\ref{sec:ranking-and-utility}
and report paired changes from All-token by model and dataset
(Tables~\ref{tab:masking-by-model} and~\ref{tab:masking-by-dataset}).

Across the four metrics, First-piece improves within-question AUROC in only
6--8 of 15 settings, and excluding Digit and Others improves it in 8--9.
Overall gains range from $0.0013$ to $0.0031$ and from $0.0054$ to $0.0083$,
respectively. First-piece improves within-question AUROC on both AIME
datasets but reduces it on GPQA-Diamond for every metric. Cross-question
effects also vary: for Gemma4-26B, excluding Digit and Others reduces the
cross-question C-Index of maximum log probability by $0.0604$
(95\% paired question-bootstrap interval: $[-0.1263,-0.0088]$).

\begin{table}[!htbp]
  \centering
  \footnotesize
  \newcommand{\maskchange}[3]{%
    \begingroup\renewcommand{\arraystretch}{0.82}%
    \begin{tabular}{@{}r@{}}#1\\[-1pt]{[#2,#3]}\end{tabular}%
    \endgroup}
  \caption{Masking effects by model. Overall gains are small, and effects vary with the metric, model, and prediction target. Model rows average the three datasets equally; Overall averages all 15 settings. All denotes the all-token baseline; $\Delta$ is the paired change from All. Brackets show 95\% paired question-bootstrap intervals. Bold marks the best condition within each row and target, including ties. Definitions of the token masks and details of the bootstrap procedure are given in Appendices~\ref{app:mask-noisy-token} and~\ref{app:diagnostic-samples}.}
  \label{tab:masking-by-model}
  \setlength{\tabcolsep}{1.2pt}
  \renewcommand{\arraystretch}{1.08}
  \begin{tabular}{@{}>{\raggedright\arraybackslash}p{.16\linewidth}*{2}{>{\centering\arraybackslash}p{.057\linewidth}*{2}{>{\raggedleft\arraybackslash}p{\dimexpr(.726\linewidth-12\tabcolsep)/4\relax}}}@{}}
    \toprule
    \multirow{3}{*}{\textbf{Model}}
    & \multicolumn{3}{c}{\textbf{within-question}}
    & \multicolumn{3}{c}{\textbf{cross-question}} \\
    & \multicolumn{3}{c}{AUROC}
    & \multicolumn{3}{c}{C-Index} \\
    \cmidrule(lr){2-4}\cmidrule(l){5-7}
    & All & $\Delta$ First-piece & \shortstack{$\Delta$ Exclude\\Digit and Others}
    & All & $\Delta$ First-piece & \shortstack{$\Delta$ Exclude\\Digit and Others} \\
    \midrule
    \multicolumn{7}{@{}l@{}}{\textbf{A. Logit gap ($\Delta_{19}$)}} \\
    \addlinespace[1pt]
    \qwenicon\hspace{0.20em}Qwen3-8B & 0.674 & \maskchange{+0.0024}{\textminus 0.0030}{0.0086} & \maskchange{\textbf{+0.0074}}{\textminus 0.0005}{0.0172} & 0.764 & \maskchange{+0.0175}{0.0010}{0.0400} & \maskchange{\textbf{+0.0294}}{0.0102}{0.0539} \\
    \qwenicon\hspace{0.20em}Qwen3-32B & \textbf{0.647} & \maskchange{\textminus 0.0020}{\textminus 0.0078}{0.0035} & \maskchange{\textminus 0.0033}{\textminus 0.0133}{0.0056} & 0.758 & \maskchange{\textbf{+0.0024}}{\textminus 0.0161}{0.0206} & \maskchange{\textminus 0.0023}{\textminus 0.0319}{0.0248} \\
    \gemmaicon\hspace{0.20em}Gemma 4-E4B & 0.656 & \maskchange{\textbf{+0.0015}}{\textminus 0.0028}{0.0063} & \maskchange{+0.0003}{\textminus 0.0095}{0.0102} & 0.776 & \maskchange{+0.0034}{\textminus 0.0096}{0.0181} & \maskchange{\textbf{+0.0123}}{\textminus 0.0022}{0.0286} \\
    \gemmaicon\hspace{0.20em}Gemma4-26B & 0.568 & \maskchange{+0.0110}{0.0020}{0.0207} & \maskchange{\textbf{+0.0219}}{\textminus 0.0019}{0.0465} & \textbf{0.732} & \maskchange{\textminus 0.0003}{\textminus 0.0217}{0.0209} & \maskchange{\textminus 0.0106}{\textminus 0.0422}{0.0139} \\
    \openaiicon\hspace{0.20em}GPT-OSS-20B & 0.529 & \maskchange{+0.0027}{\textminus 0.0021}{0.0077} & \maskchange{\textbf{+0.0151}}{0.0039}{0.0274} & 0.668 & \maskchange{\textminus 0.0003}{\textminus 0.0140}{0.0111} & \maskchange{\textbf{+0.0230}}{0.0010}{0.0494} \\
    \cmidrule(l){1-7}
    \textbf{Overall} & 0.615 & \maskchange{+0.0031}{0.0000}{0.0064} & \maskchange{\textbf{+0.0083}}{0.0012}{0.0154} & 0.740 & \maskchange{+0.0045}{\textminus 0.0032}{0.0135} & \maskchange{\textbf{+0.0104}}{\textminus 0.0013}{0.0229} \\
    \midrule
    \multicolumn{7}{@{}l@{}}{\textbf{B. Self-certainty}} \\
    \addlinespace[1pt]
    \qwenicon\hspace{0.20em}Qwen3-8B & 0.672 & \maskchange{+0.0013}{\textminus 0.0046}{0.0078} & \maskchange{\textbf{+0.0048}}{\textminus 0.0045}{0.0150} & 0.752 & \maskchange{+0.0218}{0.0026}{0.0454} & \maskchange{\textbf{+0.0330}}{0.0097}{0.0614} \\
    \qwenicon\hspace{0.20em}Qwen3-32B & \textbf{0.644} & \maskchange{\textminus 0.0035}{\textminus 0.0099}{0.0025} & \maskchange{\textminus 0.0054}{\textminus 0.0168}{0.0047} & 0.750 & \maskchange{+0.0032}{\textminus 0.0170}{0.0248} & \maskchange{\textbf{+0.0061}}{\textminus 0.0229}{0.0377} \\
    \gemmaicon\hspace{0.20em}Gemma 4-E4B & \textbf{0.655} & \maskchange{\textminus 0.0006}{\textminus 0.0078}{0.0056} & \maskchange{\textminus 0.0020}{\textminus 0.0117}{0.0070} & 0.775 & \maskchange{+0.0087}{\textminus 0.0058}{0.0278} & \maskchange{\textbf{+0.0125}}{\textminus 0.0039}{0.0323} \\
    \gemmaicon\hspace{0.20em}Gemma4-26B & 0.573 & \maskchange{+0.0087}{\textminus 0.0008}{0.0198} & \maskchange{\textbf{+0.0150}}{\textminus 0.0085}{0.0407} & \textbf{0.728} & \maskchange{\textminus 0.0048}{\textminus 0.0303}{0.0170} & \maskchange{\textminus 0.0100}{\textminus 0.0425}{0.0167} \\
    \openaiicon\hspace{0.20em}GPT-OSS-20B & 0.529 & \maskchange{+0.0021}{\textminus 0.0031}{0.0076} & \maskchange{\textbf{+0.0146}}{0.0030}{0.0275} & 0.662 & \maskchange{+0.0019}{\textminus 0.0086}{0.0129} & \maskchange{\textbf{+0.0255}}{0.0051}{0.0507} \\
    \cmidrule(l){1-7}
    \textbf{Overall} & 0.615 & \maskchange{+0.0016}{\textminus 0.0020}{0.0053} & \maskchange{\textbf{+0.0054}}{\textminus 0.0022}{0.0128} & 0.733 & \maskchange{+0.0062}{\textminus 0.0028}{0.0164} & \maskchange{\textbf{+0.0134}}{0.0007}{0.0274} \\
    \midrule
    \multicolumn{7}{@{}l@{}}{\textbf{C. Negative entropy}} \\
    \addlinespace[1pt]
    \qwenicon\hspace{0.20em}Qwen3-8B & 0.499 & \maskchange{+0.0033}{\textminus 0.0053}{0.0132} & \maskchange{\textbf{+0.0095}}{\textminus 0.0054}{0.0256} & 0.647 & \maskchange{\textbf{+0.0270}}{\textminus 0.0033}{0.0567} & \maskchange{+0.0240}{\textminus 0.0068}{0.0620} \\
    \qwenicon\hspace{0.20em}Qwen3-32B & 0.502 & \maskchange{+0.0014}{\textminus 0.0045}{0.0077} & \maskchange{\textbf{+0.0085}}{\textminus 0.0049}{0.0237} & 0.643 & \maskchange{+0.0182}{\textminus 0.0075}{0.0478} & \maskchange{\textbf{+0.0252}}{\textminus 0.0070}{0.0618} \\
    \gemmaicon\hspace{0.20em}Gemma 4-E4B & 0.666 & \maskchange{\textbf{+0.0002}}{\textminus 0.0040}{0.0048} & \maskchange{\textminus 0.0002}{\textminus 0.0071}{0.0067} & 0.774 & \maskchange{+0.0021}{\textminus 0.0065}{0.0110} & \maskchange{\textbf{+0.0054}}{\textminus 0.0070}{0.0187} \\
    \gemmaicon\hspace{0.20em}Gemma4-26B & 0.482 & \maskchange{+0.0009}{\textminus 0.0072}{0.0091} & \maskchange{\textbf{+0.0097}}{\textminus 0.0108}{0.0319} & \textbf{0.623} & \maskchange{\textminus 0.0226}{\textminus 0.0636}{0.0077} & \maskchange{\textminus 0.0541}{\textminus 0.1158}{\textminus 0.0053} \\
    \openaiicon\hspace{0.20em}GPT-OSS-20B & 0.510 & \maskchange{+0.0007}{\textminus 0.0029}{0.0041} & \maskchange{\textbf{+0.0116}}{0.0024}{0.0203} & 0.632 & \maskchange{+0.0064}{\textminus 0.0033}{0.0172} & \maskchange{\textbf{+0.0152}}{\textminus 0.0038}{0.0344} \\
    \cmidrule(l){1-7}
    \textbf{Overall} & 0.532 & \maskchange{+0.0013}{\textminus 0.0021}{0.0047} & \maskchange{\textbf{+0.0078}}{0.0009}{0.0153} & 0.664 & \maskchange{\textbf{+0.0062}}{\textminus 0.0068}{0.0190} & \maskchange{+0.0031}{\textminus 0.0143}{0.0218} \\
    \midrule
    \multicolumn{7}{@{}l@{}}{\textbf{D. Maximum log probability}} \\
    \addlinespace[1pt]
    \qwenicon\hspace{0.20em}Qwen3-8B & 0.493 & \maskchange{+0.0033}{\textminus 0.0055}{0.0128} & \maskchange{\textbf{+0.0093}}{\textminus 0.0054}{0.0256} & 0.640 & \maskchange{\textbf{+0.0260}}{\textminus 0.0034}{0.0561} & \maskchange{+0.0250}{\textminus 0.0079}{0.0641} \\
    \qwenicon\hspace{0.20em}Qwen3-32B & 0.501 & \maskchange{+0.0013}{\textminus 0.0045}{0.0074} & \maskchange{\textbf{+0.0077}}{\textminus 0.0054}{0.0224} & 0.642 & \maskchange{+0.0146}{\textminus 0.0096}{0.0411} & \maskchange{\textbf{+0.0212}}{\textminus 0.0130}{0.0595} \\
    \gemmaicon\hspace{0.20em}Gemma 4-E4B & 0.667 & \maskchange{\textbf{+0.0010}}{\textminus 0.0032}{0.0055} & \maskchange{\textminus 0.0008}{\textminus 0.0087}{0.0064} & 0.775 & \maskchange{+0.0016}{\textminus 0.0087}{0.0113} & \maskchange{\textbf{+0.0060}}{\textminus 0.0061}{0.0195} \\
    \gemmaicon\hspace{0.20em}Gemma4-26B & 0.471 & \maskchange{+0.0038}{\textminus 0.0055}{0.0138} & \maskchange{\textbf{+0.0086}}{\textminus 0.0115}{0.0284} & \textbf{0.610} & \maskchange{\textminus 0.0325}{\textminus 0.0693}{\textminus 0.0032} & \maskchange{\textminus 0.0604}{\textminus 0.1263}{\textminus 0.0088} \\
    \openaiicon\hspace{0.20em}GPT-OSS-20B & 0.508 & \maskchange{+0.0001}{\textminus 0.0036}{0.0038} & \maskchange{\textbf{+0.0128}}{0.0044}{0.0209} & 0.628 & \maskchange{+0.0052}{\textminus 0.0046}{0.0148} & \maskchange{\textbf{+0.0102}}{\textminus 0.0055}{0.0277} \\
    \cmidrule(l){1-7}
    \textbf{Overall} & 0.528 & \maskchange{+0.0019}{\textminus 0.0015}{0.0056} & \maskchange{\textbf{+0.0075}}{0.0006}{0.0147} & 0.659 & \maskchange{\textbf{+0.0030}}{\textminus 0.0088}{0.0151} & \maskchange{+0.0004}{\textminus 0.0171}{0.0197} \\
    \bottomrule
  \end{tabular}
\end{table}

\begin{table}[!htbp]
  \centering
  \footnotesize
  \newcommand{\maskchange}[3]{%
    \begingroup\renewcommand{\arraystretch}{0.9}%
    \begin{tabular}{@{}r@{}}#1\\[-1pt]{[#2,#3]}\end{tabular}%
    \endgroup}
  \caption{Masking effects by dataset. First-piece improves within-question AUROC on both AIME datasets but reduces it on GPQA-Diamond for all four metrics. Rows average the five models equally. All denotes the all-token baseline; $\Delta$ is the paired change from All. Brackets show 95\% paired question-bootstrap intervals. Bold marks the best condition within each row and target, including ties. Definitions of the token masks and details of the bootstrap procedure are given in Appendices~\ref{app:mask-noisy-token} and~\ref{app:diagnostic-samples}.}
  \label{tab:masking-by-dataset}
  \setlength{\tabcolsep}{1.2pt}
  \renewcommand{\arraystretch}{1.08}
  \begin{tabular}{@{}>{\raggedright\arraybackslash}p{.16\linewidth}*{2}{>{\centering\arraybackslash}p{.057\linewidth}*{2}{>{\raggedleft\arraybackslash}p{\dimexpr(.726\linewidth-12\tabcolsep)/4\relax}}}@{}}
    \toprule
    \multirow{3}{*}{\textbf{Dataset}}
    & \multicolumn{3}{c}{\textbf{within-question}}
    & \multicolumn{3}{c}{\textbf{cross-question}} \\
    & \multicolumn{3}{c}{AUROC}
    & \multicolumn{3}{c}{C-Index} \\
    \cmidrule(lr){2-4}\cmidrule(l){5-7}
    & All & $\Delta$ First-piece & \shortstack{$\Delta$ Exclude\\Digit and Others}
    & All & $\Delta$ First-piece & \shortstack{$\Delta$ Exclude\\Digit and Others} \\
    \midrule
    \multicolumn{7}{@{}l@{}}{\textbf{A. Logit gap ($\Delta_{19}$)}} \\
    \addlinespace[1pt]
    AIME 2024 & 0.624 & \maskchange{+0.0075}{0.0015}{0.0139} & \maskchange{\textbf{+0.0247}}{0.0076}{0.0426} & 0.759 & \maskchange{+0.0044}{\textminus 0.0106}{0.0228} & \maskchange{\textbf{+0.0161}}{\textminus 0.0121}{0.0461} \\
    AIME 2025 & 0.637 & \maskchange{\textbf{+0.0052}}{\textminus 0.0006}{0.0116} & \maskchange{+0.0006}{\textminus 0.0096}{0.0109} & 0.783 & \maskchange{\textbf{+0.0058}}{\textminus 0.0104}{0.0252} & \maskchange{+0.0057}{\textminus 0.0147}{0.0275} \\
    GPQA-Diamond & \textbf{0.583} & \maskchange{\textminus 0.0034}{\textminus 0.0072}{0.0002} & \maskchange{\textminus 0.0004}{\textminus 0.0040}{0.0037} & 0.677 & \maskchange{+0.0034}{\textminus 0.0029}{0.0101} & \maskchange{\textbf{+0.0092}}{0.0017}{0.0166} \\
    \midrule
    \multicolumn{7}{@{}l@{}}{\textbf{B. Self-certainty}} \\
    \addlinespace[1pt]
    AIME 2024 & 0.623 & \maskchange{+0.0070}{\textminus 0.0008}{0.0146} & \maskchange{\textbf{+0.0216}}{0.0027}{0.0416} & 0.749 & \maskchange{+0.0138}{\textminus 0.0047}{0.0371} & \maskchange{\textbf{+0.0250}}{\textminus 0.0060}{0.0605} \\
    AIME 2025 & 0.638 & \maskchange{\textbf{+0.0023}}{\textminus 0.0042}{0.0087} & \maskchange{\textminus 0.0049}{\textminus 0.0167}{0.0052} & 0.778 & \maskchange{+0.0006}{\textminus 0.0191}{0.0221} & \maskchange{\textbf{+0.0056}}{\textminus 0.0183}{0.0314} \\
    GPQA-Diamond & \textbf{0.583} & \maskchange{\textminus 0.0044}{\textminus 0.0083}{\textminus 0.0005} & \maskchange{\textminus 0.0004}{\textminus 0.0040}{0.0035} & 0.673 & \maskchange{+0.0041}{\textminus 0.0027}{0.0108} & \maskchange{\textbf{+0.0096}}{0.0016}{0.0177} \\
    \midrule
    \multicolumn{7}{@{}l@{}}{\textbf{C. Negative entropy}} \\
    \addlinespace[1pt]
    AIME 2024 & 0.522 & \maskchange{+0.0054}{\textminus 0.0005}{0.0120} & \maskchange{\textbf{+0.0257}}{0.0093}{0.0437} & 0.692 & \maskchange{\textbf{+0.0073}}{\textminus 0.0208}{0.0372} & \maskchange{\textminus 0.0131}{\textminus 0.0534}{0.0301} \\
    AIME 2025 & 0.526 & \maskchange{\textbf{+0.0029}}{\textminus 0.0045}{0.0108} & \maskchange{\textminus 0.0011}{\textminus 0.0140}{0.0120} & 0.657 & \maskchange{+0.0021}{\textminus 0.0244}{0.0267} & \maskchange{\textbf{+0.0159}}{\textminus 0.0186}{0.0535} \\
    GPQA-Diamond & \textbf{0.547} & \maskchange{\textminus 0.0044}{\textminus 0.0076}{\textminus 0.0013} & \maskchange{\textminus 0.0012}{\textminus 0.0040}{0.0018} & 0.641 & \maskchange{\textbf{+0.0092}}{0.0030}{0.0162} & \maskchange{+0.0066}{\textminus 0.0021}{0.0151} \\
    \midrule
    \multicolumn{7}{@{}l@{}}{\textbf{D. Maximum log probability}} \\
    \addlinespace[1pt]
    AIME 2024 & 0.518 & \maskchange{+0.0077}{0.0012}{0.0156} & \maskchange{\textbf{+0.0264}}{0.0099}{0.0443} & 0.685 & \maskchange{\textbf{+0.0019}}{\textminus 0.0228}{0.0287} & \maskchange{\textminus 0.0164}{\textminus 0.0575}{0.0291} \\
    AIME 2025 & 0.520 & \maskchange{\textbf{+0.0027}}{\textminus 0.0041}{0.0104} & \maskchange{\textminus 0.0031}{\textminus 0.0163}{0.0102} & 0.652 & \maskchange{\textminus 0.0025}{\textminus 0.0291}{0.0218} & \maskchange{\textbf{+0.0114}}{\textminus 0.0230}{0.0475} \\
    GPQA-Diamond & \textbf{0.546} & \maskchange{\textminus 0.0047}{\textminus 0.0077}{\textminus 0.0018} & \maskchange{\textminus 0.0008}{\textminus 0.0037}{0.0023} & 0.639 & \maskchange{\textbf{+0.0095}}{0.0030}{0.0166} & \maskchange{+0.0061}{\textminus 0.0027}{0.0150} \\
    \bottomrule
  \end{tabular}
\end{table}

\FloatBarrier
\section{Details for Section 5}
\label{app:signal-localization-details}

Table~\ref{tab:temporal-localization-by-model} compares model-level results for 1,000-token prefixes and suffixes with full-trace scores.

\begin{table}[!htbp]
  \centering
  \footnotesize
  \caption{Temporal localization across models. Both prediction targets use
  the same three aggregation regions: the first 1,000 tokens, the final 1,000
  tokens, and the full trace. The cross-question C-Index ranks empirical pass rates
  across questions, whereas within-question AUROC ranks correct and incorrect
  traces for the same question. Each model row equally averages AIME 2024,
  AIME 2025, and GPQA-Diamond; Overall equally averages all model--dataset
  combinations. Higher is better. Blue shading marks the selected windows
  (cross-question prefix and within-question suffix), not row maxima.}
  \label{tab:temporal-localization-by-model}
  \setlength{\tabcolsep}{4.4pt}
  \renewcommand{\arraystretch}{1.08}
  \begin{tabular}{@{}l@{\hspace{9pt}}ccc@{\hspace{12pt}}ccc@{}}
    \toprule
    \multirow{2}{*}{\textbf{Model}}
    & \multicolumn{3}{c}{\textbf{cross-question C-Index}}
    & \multicolumn{3}{c}{\textbf{within-question AUROC}} \\
    \cmidrule(lr){2-4}\cmidrule(l){5-7}
    & \rankdefault{Prefix (1k)} & Suffix (1k) & Full Trace
    & Prefix (1k) & \rankdefault{Suffix (1k)} & Full Trace \\
    \midrule
    \qwenicon\hspace{0.20em}Qwen3-8B
      & \rankdefault{0.709} & 0.774 & 0.764
      & 0.505 & \rankdefault{0.683} & 0.674 \\
    \qwenicon\hspace{0.20em}Qwen3-32B
      & \rankdefault{0.728} & 0.769 & 0.758
      & 0.502 & \rankdefault{0.652} & 0.647 \\
    \gemmaicon\hspace{0.20em}Gemma4-4B
      & \rankdefault{0.721} & 0.748 & 0.776
      & 0.489 & \rankdefault{0.675} & 0.656 \\
    \gemmaicon\hspace{0.20em}Gemma4-26B
      & \rankdefault{0.753} & 0.701 & 0.732
      & 0.514 & \rankdefault{0.533} & 0.568 \\
    \openaiicon\hspace{0.20em}GPT-OSS-20B
      & \rankdefault{0.701} & 0.664 & 0.668
      & 0.491 & \rankdefault{0.584} & 0.529 \\
    \midrule
    \textbf{Overall}
      & \rankdefault{0.723} & \textbf{0.731} & \textbf{0.740}
      & \textbf{0.500} & \rankdefault{0.626} & \textbf{0.615} \\
    \bottomrule
  \end{tabular}
\end{table}

\subsection{Sensitivity to Window Length}

Section~\ref{sec:signal-localization} uses matched 1,000-token windows to
compare where evidence for the two prediction targets becomes available. Here
we test whether the results depend on window length, the summary and
final-answer tokens included in late windows, or the choice of certainty metric.
We retain the main text's cross-question C-Index and within-question AUROC;
the window-length comparisons focus on prefixes for cross-question difficulty
and suffixes for within-question trace correctness.

We use a fixed 1,000-token window across models and datasets. Table~\ref{tab:temporal-robustness} shows that early
prefixes remain informative about cross-question difficulty and short suffixes
retain higher within-question AUROC than the 30,000-token reference across
neighboring window lengths. From 0.5k to 2k tokens, prefix
cross-question C-Index ranges from $0.691$ to $0.735$, while suffix
within-question AUROC ranges from $0.624$ to $0.638$; every suffix value exceeds
the $0.611$ reference obtained from the final 30,000 tokens. The cross-question C-Index generally increases as
the prefix expands. The within-question AUROC is strongest for shorter suffixes and
generally decreases as the window expands to include earlier trace regions.
Discrimination varies across models and datasets.

\begin{table}[!htbp]
  \centering
  \footnotesize
  \caption{Effect of absolute window length on prefix cross-question C-Index
  and suffix within-question AUROC. Entries give equal-weight averages across the 15
  model--dataset combinations, with the observed minimum--maximum range
  in brackets. Windows count generated tokens. The 30,000-token suffix approximates
  the full trace, covering 98.4\% of responses. Higher is better.}
  \label{tab:temporal-robustness}
  \newcommand{\temporalrange}[3]{\shortstack{$#1$\\[-1pt]{\scriptsize$[#2,#3]$}}}
  \setlength{\tabcolsep}{5.0pt}
  \renewcommand{\arraystretch}{1.04}
  \begin{tabular*}{\linewidth}{@{\extracolsep{\fill}}lccccc@{}}
    \toprule
    \multicolumn{6}{l}{\textbf{(a) 100--500 generated tokens}} \\
    \addlinespace[2pt]
    Readout & 100 & 200 & 300 & 400 & 500 \\
    \midrule
    cross-question: prefix C-Index
      & \temporalrange{0.578}{0.449}{0.664}
      & \temporalrange{0.642}{0.492}{0.731}
      & \temporalrange{0.670}{0.550}{0.744}
      & \temporalrange{0.682}{0.595}{0.754}
      & \temporalrange{0.691}{0.608}{0.754} \\
    within-question: suffix AUROC
      & \temporalrange{0.639}{0.501}{0.742}
      & \temporalrange{0.639}{0.402}{0.756}
      & \temporalrange{0.640}{0.443}{0.758}
      & \temporalrange{0.633}{0.414}{0.771}
      & \temporalrange{0.638}{0.429}{0.791} \\
    \bottomrule
  \end{tabular*}

  \vspace{5pt}

  \begin{tabular*}{\linewidth}{@{\extracolsep{\fill}}lccccc@{}}
    \toprule
    \multicolumn{6}{l}{\textbf{(b) 1,000--5,000 generated tokens}} \\
    \addlinespace[2pt]
    Readout & 1,000 & 2,000 & 3,000 & 4,000 & 5,000 \\
    \midrule
    cross-question: prefix C-Index
      & \temporalrange{0.723}{0.637}{0.808}
      & \temporalrange{0.735}{0.649}{0.841}
      & \temporalrange{0.737}{0.652}{0.857}
      & \temporalrange{0.731}{0.651}{0.841}
      & \temporalrange{0.726}{0.649}{0.846} \\
    within-question: suffix AUROC
      & \temporalrange{0.626}{0.509}{0.737}
      & \temporalrange{0.624}{0.448}{0.738}
      & \temporalrange{0.610}{0.376}{0.732}
      & \temporalrange{0.611}{0.408}{0.740}
      & \temporalrange{0.606}{0.411}{0.721} \\
    \bottomrule
  \end{tabular*}

  \vspace{5pt}

  \begin{tabular*}{\linewidth}{@{\extracolsep{\fill}}lcccc@{}}
    \toprule
    \multicolumn{5}{l}{\textbf{(c) 10,000--30,000 generated tokens}} \\
    \addlinespace[2pt]
    Readout & 10,000 & 15,000 & 20,000 & 30,000 \\
    \midrule
    cross-question: prefix C-Index
      & \temporalrange{0.720}{0.642}{0.870}
      & \temporalrange{0.731}{0.645}{0.875}
      & \temporalrange{0.740}{0.645}{0.875}
      & \temporalrange{0.742}{0.638}{0.875} \\
    within-question: suffix AUROC
      & \temporalrange{0.609}{0.469}{0.713}
      & \temporalrange{0.611}{0.461}{0.713}
      & \temporalrange{0.611}{0.461}{0.705}
      & \temporalrange{0.611}{0.480}{0.706} \\
    \bottomrule
  \end{tabular*}
\end{table}

Table~\ref{tab:temporal-robustness-relative} repeats the sweep using fractions
of each trace. Prefix cross-question performance again generally rises as more
context is included. Suffix within-question AUROC instead peaks at $0.651$ for
the final $5\%$ and falls to $0.616$--$0.619$ over the final $20\%$--$70\%$.
The broad temporal pattern also appears with relative windows, although
discrimination varies across settings and window lengths.

\begin{table}[!htbp]
  \centering
  \footnotesize
  \caption{Effect of relative window length on prefix cross-question C-Index
  and suffix within-question AUROC. Entries average the 15 model--dataset
  combinations equally, with the observed minimum--maximum range in
  brackets. Windows are fractions of each trace. Higher is better.}
  \label{tab:temporal-robustness-relative}
  \newcommand{\temporalrelrange}[3]{\shortstack{$#1$\\[-1pt]{\scriptsize$[#2,#3]$}}}
  \setlength{\tabcolsep}{5.0pt}
  \renewcommand{\arraystretch}{1.04}
  \begin{tabular*}{\linewidth}{@{\extracolsep{\fill}}lccccc@{}}
    \toprule
    \multicolumn{6}{l}{\textbf{(a) 0.1\%--5\% of the trace}} \\
    \addlinespace[2pt]
    Readout & 0.1\% & 0.5\% & 1\% & 2\% & 5\% \\
    \midrule
    cross-question: prefix C-Index
      & \temporalrelrange{0.508}{0.210}{0.851}
      & \temporalrelrange{0.627}{0.493}{0.831}
      & \temporalrelrange{0.672}{0.542}{0.794}
      & \temporalrelrange{0.693}{0.565}{0.797}
      & \temporalrelrange{0.688}{0.605}{0.789} \\
    within-question: suffix AUROC
      & \temporalrelrange{0.611}{0.447}{0.780}
      & \temporalrelrange{0.621}{0.457}{0.753}
      & \temporalrelrange{0.628}{0.497}{0.736}
      & \temporalrelrange{0.636}{0.501}{0.755}
      & \temporalrelrange{0.651}{0.567}{0.743} \\
    \bottomrule
  \end{tabular*}

  \vspace{5pt}

  \begin{tabular*}{\linewidth}{@{\extracolsep{\fill}}lccccc@{}}
    \toprule
    \multicolumn{6}{l}{\textbf{(b) 10\%--70\% of the trace}} \\
    \addlinespace[2pt]
    Readout & 10\% & 20\% & 30\% & 50\% & 70\% \\
    \midrule
    cross-question: prefix C-Index
      & \temporalrelrange{0.697}{0.590}{0.819}
      & \temporalrelrange{0.708}{0.564}{0.852}
      & \temporalrelrange{0.715}{0.564}{0.874}
      & \temporalrelrange{0.719}{0.533}{0.846}
      & \temporalrelrange{0.724}{0.540}{0.873} \\
    within-question: suffix AUROC
      & \temporalrelrange{0.636}{0.505}{0.745}
      & \temporalrelrange{0.617}{0.445}{0.720}
      & \temporalrelrange{0.616}{0.469}{0.718}
      & \temporalrelrange{0.619}{0.490}{0.715}
      & \temporalrelrange{0.618}{0.479}{0.713} \\
    \bottomrule
  \end{tabular*}
\end{table}

\subsection{Effects of Summary and Answer Content}

Late windows overlap summaries and final answers, so their higher
within-question scores could reflect which parts of a response they contain.
We therefore compare the suffix before and after removing final-answer spans
and compare it with the full summary. We also divide the summary into four consecutive
parts, labeled Q1--Q4 from first to last, each containing approximately one
quarter of the summary tokens.
Table~\ref{tab:segment-composition-controls} reports the absolute scores
under these controls.

\begin{table}[!htbp]
  \centering
  \footnotesize
  \caption{Scores for the final 1,000-token suffix before and after removing
  final-answer spans, the full summary, and its four consecutive quarters
  (Q1--Q4). Quarters contain approximately equal token counts and are ordered
  from first to last. Values average the 15 model--dataset settings equally;
  higher is better.}
  \label{tab:segment-composition-controls}
  \setlength{\tabcolsep}{3.0pt}
  \renewcommand{\arraystretch}{1.05}
  \begin{tabular}{@{}llrr@{}}
    \toprule
    Control dimension & Aggregation region
      & within-question AUROC & cross-question C-Index \\
    \midrule
    \multirow{2}{*}{Answer-span content}
      & Suffix (1k) & 0.6260 & 0.7310 \\
      & Suffix (1k), answers removed & 0.6279 & 0.7334 \\
    \addlinespace[2pt]
    \multirow{2}{*}{Summary extent}
      & Full summary & 0.6100 & 0.7414 \\
      & Suffix (1k) & 0.6260 & 0.7310 \\
    \addlinespace[2pt]
    \multirow{4}{*}{Within-summary position}
      & Summary Q1 & 0.5371 & 0.6950 \\
      & Summary Q2 & 0.5353 & 0.7021 \\
      & Summary Q3 & 0.5684 & 0.7090 \\
      & Summary Q4 & 0.6208 & 0.6822 \\
    \bottomrule
  \end{tabular}
\end{table}

Answer removal leaves both aggregate scores nearly unchanged. The within-question
AUROC is higher for the suffix than for the full summary ($0.6260$ versus
$0.6100$) and rises from $0.5371$ in summary Q1 to $0.6208$ in Q4. The
cross-question scores do not show the same increase toward the end.
Summary content changes from Q1 to Q4. These comparisons do not isolate the
effect of position from changes in content.

\subsection{Comparisons across Certainty Metrics}
\label{app:temporal-metric-comparisons}

\begin{table}[!htbp]
  \centering
  \footnotesize
  \caption{Full-trace, prefix, and suffix scores across four certainty metrics. Both
  panels report equal-weight averages over the same 15 model--dataset
  combinations. For each metric, panel (a) compares the cross-question C-Index
  of the full trace with that of the 1,000-token prefix, and panel (b) compares
  the within-question AUROC of the full trace with that of the 1,000-token suffix. Each
  difference is the prefix or suffix score minus the full-trace score; the
  final column counts combinations on which that prefix or suffix score is higher.
  No exact ties occur. Blue shading identifies the prefix in panel (a) and
  the suffix in panel (b) and
  does not denote the largest value in a row. Higher is better.}
  \label{tab:temporal-metric-replication}
  \setlength{\tabcolsep}{6.0pt}
  \renewcommand{\arraystretch}{1.07}
  \begin{tabular}{@{}lrrrr@{}}
    \toprule
    \multicolumn{5}{l}{\textbf{(a) cross-question C-Index}} \\
    \addlinespace[2pt]
    Metric & Full trace & \rankdefault{Prefix (1k)}
      & \shortstack{Prefix (1k)\\$-$ full trace}
      & \shortstack{Combinations with\\prefix $>$ full trace} \\
    \midrule
    Logit gap & 0.740 & \rankdefault{0.723} & $-0.017$ & 8/15 \\
    Self-certainty & 0.733 & \rankdefault{0.717} & $-0.016$ & 7/15 \\
    Negative entropy & 0.664 & \rankdefault{0.708} & $+0.045$ & 12/15 \\
    Maximum log probability & 0.659 & \rankdefault{0.709} & $+0.050$ & 12/15 \\
    \bottomrule
  \end{tabular}

  \vspace{5pt}

  \begin{tabular}{@{}lrrrr@{}}
    \toprule
    \multicolumn{5}{l}{\textbf{(b) within-question AUROC}} \\
    \addlinespace[2pt]
    Metric & Full trace & \rankdefault{Suffix (1k)}
      & \shortstack{Suffix (1k)\\$-$ full trace}
      & \shortstack{Combinations with\\suffix $>$ full trace} \\
    \midrule
    Logit gap & 0.615 & \rankdefault{0.626} & $+0.011$ & 9/15 \\
    Self-certainty & 0.615 & \rankdefault{0.626} & $+0.012$ & 9/15 \\
    Negative entropy & 0.532 & \rankdefault{0.589} & $+0.057$ & 14/15 \\
    Maximum log probability & 0.528 & \rankdefault{0.585} & $+0.057$ & 13/15 \\
    \bottomrule
  \end{tabular}
\end{table}

Finally, we check whether the early cross-question signal and the higher
within-question scores in suffixes also appear with metrics other than
logit gap. In Table~\ref{tab:temporal-metric-replication},
we apply the same full-trace and 1,000-token comparisons to all four metrics
defined in Sections~\ref{section:preliminaries} and~\ref{sec:evaluation-framework}.

Across all four metrics, suffixes outperform full traces in average
within-question AUROC. Prefix cross-question scores retain most full-trace
performance for logit gap and self-certainty and improve it for negative
entropy and maximum log probability. The latter uses only the top token. Results vary across settings.
Section~\ref{sec:test-time-compute} evaluates allocation from early scores
and voting from late scores.

\FloatBarrier
\section{Details for Section 6}
\label{app:test-time-compute-certainty}

This appendix specifies the TACS configuration in
Table~\ref{tab:tts-component-ablation} and reports its results by model and dataset.
We then examine how TACS distributes samples across questions
and how often sampling reaches the trace limit, before comparing it with
fixed-budget and adaptive baselines.

\subsection{Experimental Protocol}
\label{app:tts-protocol}

We evaluate five models on 30 questions from each AIME dataset and 198
from GPQA-Diamond. TACS uses the top-1--top-20 logit gap $\Delta_{19}$
(Equation~\ref{eq:ranked-logit-gap}), averaged over the first 1,000 tokens
for allocation and the final 500 tokens for voting. For traces shorter than
the scoring window, we use their full length. Appendix~\ref{app:tacs-simulation} provides sampling
and cost-accounting details.

We calibrate the model-specific allocation thresholds on BRUMO 2025 \citep{Balunovic2025Matharena}.
The allocation thresholds $\theta_m$ are 19.0 for Qwen3-8B, 16.5 for
Qwen3-32B, 14.0 for Gemma 4-E4B, 16.5 for Gemma4-26B, and 10.0 for
GPT-OSS-20B. Each model's threshold is shared across datasets. We generate complete traces
until 32 have prefix scores $\geq\theta_m$ or 256 have been sampled.

We use rank-weighted voting for observed responses, including those below
the prefix threshold. For $n$ voting responses, let $r_i$ be response $i$'s
ascending certainty rank from 1 to $n$, with tied scores assigned their
average rank. Its vote weight is $w_i=r_i^p$, with rank exponent $p=1$.
The answer with the largest total rank weight wins.
The fixed policy uses certainty scores and generated answers to stop and vote.

\subsection{Detailed Results and Voting Comparisons}

Table~\ref{tab:tts-model-dataset-results} reports the main comparison across
15 model--dataset settings. Within either sampling condition, all four
voting methods observe the same traces, so their accuracy differences arise
from answer aggregation. Under early allocation, Suffix 500 rank-weighted
voting achieves the highest accuracy in all nine settings involving Qwen or
Gemma 4-E4B. For Gemma4-26B and GPT-OSS-20B, the voting method with the
highest accuracy varies across datasets.

Sampling changes the candidates. With Suffix 500 rank-weighted voting,
early allocation raises Qwen3-32B's AIME 2025 accuracy from $80.08\%$ to
$81.38\%$, but lowers Gemma4-26B's from $96.67\%$ to $94.92\%$.
All relative token costs are below one, ranging from $0.125$ to
$0.312$. Overall accuracy is $79.54\%$ at relative cost $0.176$; with fixed
256-trace sampling, the same voting rule yields $79.71\%$.

\begin{table}[!htbp]
  \centering
  \small
  \caption{Voting by model and dataset.
  Majority voting gives one vote per response; other columns use
  rank-weighted voting ($p=1$) with mean logit gap over the indicated region.
  Methods within each panel share traces and token cost. Overall averages
  the 15 settings equally; relative cost is each setting's total tokens divided
  by its fixed-256 total. Bold marks row maxima within each panel,
  including ties at displayed precision.}
  \label{tab:tts-model-dataset-results}
  \renewcommand{\arraystretch}{1.06}
  \setlength{\tabcolsep}{0pt}
  \begin{tabular}{@{}p{0.19\linewidth}p{0.19\linewidth}*{5}{>{\centering\arraybackslash}p{0.124\linewidth}}@{}}
    \toprule
    \multicolumn{7}{@{}l}{\textbf{(a) Fixed-budget sampling: 256 traces}} \\
    \addlinespace[2pt]
    & & \multicolumn{4}{c}{Accuracy (\%) $\uparrow$} & \shortstack{Relative\\token cost $\downarrow$} \\
    \cmidrule(lr){3-6}
    \textbf{Model} & \textbf{Dataset} & \shortstack{Majority\\voting} & \shortstack{Full\\trace} & \shortstack{Final 1,000\\tokens} & \shortstack{Final 500\\tokens} & \\
    \midrule
    \qwenicon\hspace{0.2em}Qwen3-8B & AIME 2024 & 80.00 & \textbf{83.02} & 82.94 & 82.76 & 1.000 \\
     & AIME 2025 & 79.77 & 79.35 & 79.61 & \textbf{79.79} & 1.000 \\
     & GPQA-Diamond & 66.52 & 66.07 & 66.54 & \textbf{66.81} & 1.000 \\
    \addlinespace[2pt]
    \qwenicon\hspace{0.2em}Qwen3-32B & AIME 2024 & 86.02 & 90.55 & 88.80 & \textbf{90.60} & 1.000 \\
     & AIME 2025 & 80.00 & 80.05 & \textbf{80.08} & \textbf{80.08} & 1.000 \\
     & GPQA-Diamond & 72.87 & 72.72 & \textbf{73.45} & 73.25 & 1.000 \\
    \addlinespace[2pt]
    \gemmaicon\hspace{0.2em}Gemma 4-E4B & AIME 2024 & 64.38 & 66.88 & 67.21 & \textbf{67.94} & 1.000 \\
     & AIME 2025 & 56.61 & 58.13 & 58.05 & \textbf{59.01} & 1.000 \\
     & GPQA-Diamond & 65.35 & 66.15 & 65.67 & \textbf{66.19} & 1.000 \\
    \addlinespace[2pt]
    \gemmaicon\hspace{0.2em}Gemma4-26B & AIME 2024 & \textbf{93.33} & \textbf{93.33} & \textbf{93.33} & \textbf{93.33} & 1.000 \\
     & AIME 2025 & 96.59 & \textbf{96.67} & 96.61 & \textbf{96.67} & 1.000 \\
     & GPQA-Diamond & 84.24 & \textbf{84.54} & 84.15 & 84.47 & 1.000 \\
    \addlinespace[2pt]
    \openaiicon\hspace{0.2em}GPT-OSS-20B & AIME 2024 & \textbf{93.33} & \textbf{93.33} & \textbf{93.33} & \textbf{93.33} & 1.000 \\
     & AIME 2025 & \textbf{94.01} & 93.33 & 93.46 & 93.41 & 1.000 \\
     & GPQA-Diamond & 67.61 & 67.67 & \textbf{68.10} & 68.06 & 1.000 \\
    \midrule
    \textbf{Overall} &  & 78.71 & 79.45 & 79.42 & \textbf{79.71} & 1.000 \\
    \midrule
    \addlinespace[3pt]
    \multicolumn{7}{@{}l}{\textbf{(b) Early allocation: target of 32 qualifying traces; at most 256 traces}} \\
    \addlinespace[2pt]
    & & \multicolumn{4}{c}{Accuracy (\%) $\uparrow$} & \shortstack{Relative\\token cost $\downarrow$} \\
    \cmidrule(lr){3-6}
    \textbf{Model} & \textbf{Dataset} & \shortstack{Majority\\voting} & \shortstack{Full\\trace} & \shortstack{Final 1,000\\tokens} & \shortstack{Final 500\\tokens} & \\
    \midrule
    \qwenicon\hspace{0.2em}Qwen3-8B & AIME 2024 & 80.70 & 83.31 & 83.54 & \textbf{83.62} & 0.126 \\
     & AIME 2025 & 78.36 & 77.79 & 78.96 & \textbf{79.14} & 0.125 \\
     & GPQA-Diamond & 65.80 & 66.10 & 66.14 & \textbf{66.24} & 0.174 \\
    \addlinespace[2pt]
    \qwenicon\hspace{0.2em}Qwen3-32B & AIME 2024 & 88.26 & 89.45 & 89.38 & \textbf{90.21} & 0.125 \\
     & AIME 2025 & 81.17 & 81.30 & 81.12 & \textbf{81.38} & 0.125 \\
     & GPQA-Diamond & 72.60 & 73.37 & 73.63 & \textbf{73.65} & 0.184 \\
    \addlinespace[2pt]
    \gemmaicon\hspace{0.2em}Gemma 4-E4B & AIME 2024 & 64.40 & 67.32 & 67.60 & \textbf{67.68} & 0.132 \\
     & AIME 2025 & 54.53 & 56.15 & 56.59 & \textbf{58.26} & 0.125 \\
     & GPQA-Diamond & 64.74 & 65.29 & 65.10 & \textbf{65.46} & 0.300 \\
    \addlinespace[2pt]
    \gemmaicon\hspace{0.2em}Gemma4-26B & AIME 2024 & \textbf{93.93} & 93.88 & 93.44 & 93.44 & 0.135 \\
     & AIME 2025 & 94.77 & \textbf{95.03} & 94.53 & 94.92 & 0.128 \\
     & GPQA-Diamond & 83.90 & \textbf{84.49} & 84.17 & 84.22 & 0.312 \\
    \addlinespace[2pt]
    \openaiicon\hspace{0.2em}GPT-OSS-20B & AIME 2024 & 93.31 & \textbf{93.33} & 93.15 & 93.20 & 0.279 \\
     & AIME 2025 & 93.12 & 92.76 & \textbf{93.54} & 93.49 & 0.177 \\
     & GPQA-Diamond & 66.94 & 67.34 & 67.91 & \textbf{68.20} & 0.186 \\
    \midrule
    \textbf{Overall} &  & 78.44 & 79.13 & 79.25 & \textbf{79.54} & 0.176 \\
    \bottomrule
  \end{tabular}
\end{table}

\subsection{Sampling Allocation and Stopping}
\label{app:sampling-allocation-stopping}

\begin{figure}[!htbp]
  \centering
  \includegraphics[width=0.75\linewidth]{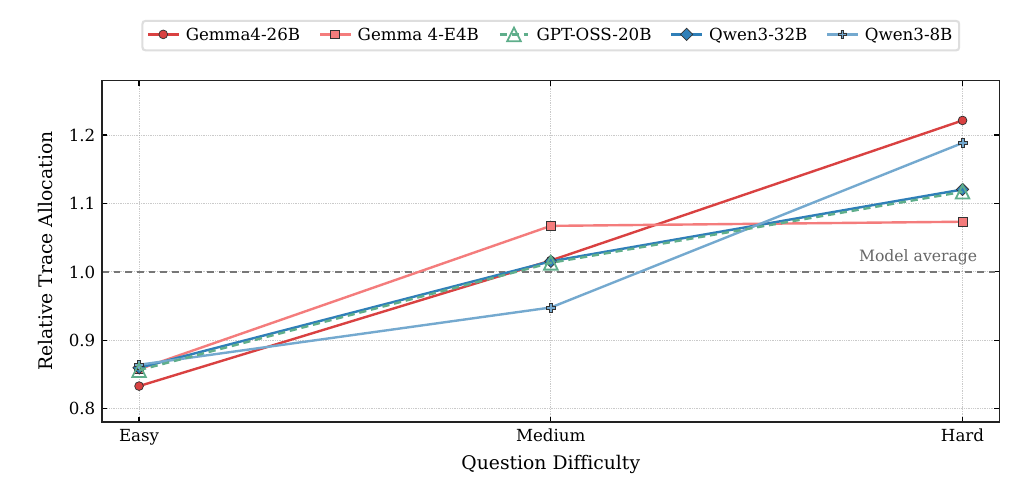}
  \caption{TACS allocates more traces to harder question groups.
  Group means are normalized by each model's mean across
  all 258 questions; the horizontal dashed line marks this reference.}
  \label{fig:sampling-allocation-by-difficulty}
\end{figure}

A shared target of 32 qualifying traces produces different sample counts
across questions. For Figure~\ref{fig:sampling-allocation-by-difficulty},
we define difficulty as one minus empirical accuracy on 512 reference
responses per model--question pair, counting invalid answers as incorrect.
Within each model, we pool the datasets and divide questions into easy,
medium, and hard groups by difficulty rank, keeping ties together.
We normalize each group's mean trace count by the model-wide mean across
all 258 questions, weighting questions equally.
Figure~\ref{fig:sampling-allocation-by-difficulty} shows that the hard group
receives more traces than the easy group for every model.

Table~\ref{tab:tts-stopping-statistics} reports trace counts: every model has a
median of 32, with means ranging from $41.14$ to $71.50$.
About $13\%$ of sampling runs reach the 256-trace limit for the Gemma
models, compared with $1.94\%$--$2.71\%$ for the other models.
Questions are weighted equally here; Table~\ref{tab:tts-adaptive-baselines}
weights datasets equally and reports a mean of 43.5 traces for TACS.

\begin{table}[!htbp]
  \centering
  \small
  \caption{Sample counts under early allocation, with a target of 32
  qualifying traces and a limit of 256. Means and medians summarize trace
  counts per sampling run across each model's 258 questions, weighted equally.
  The final column gives the percentage of runs using 256 traces.
  Table~\ref{tab:tts-adaptive-baselines} weights datasets equally.}
  \label{tab:tts-stopping-statistics}
  \renewcommand{\arraystretch}{1.06}
  \setlength{\tabcolsep}{0pt}
  \begin{tabular}{@{}p{0.36\linewidth}*{4}{>{\centering\arraybackslash}p{0.16\linewidth}}@{}}
    \toprule
    \textbf{Model} & \shortstack{Number of\\questions} & \shortstack{Mean trace\\count} & \shortstack{Median trace\\count} & \shortstack{Sampling runs\\at the trace\\limit (\%)} \\
    \midrule
    \qwenicon\hspace{0.2em}Qwen3-8B & 258 & 41.14 & 32 & 1.94 \\
    \qwenicon\hspace{0.2em}Qwen3-32B & 258 & 41.49 & 32 & 1.97 \\
    \gemmaicon\hspace{0.2em}Gemma 4-E4B & 258 & 71.50 & 32 & 12.82 \\
    \gemmaicon\hspace{0.2em}Gemma4-26B & 258 & 70.70 & 32 & 12.92 \\
    \openaiicon\hspace{0.2em}GPT-OSS-20B & 258 & 43.24 & 32 & 2.71 \\
    \bottomrule
  \end{tabular}
\end{table}

\subsection{Baseline Comparisons}
\label{app:baseline-comparisons}

Table~\ref{tab:tts-fixed-budget-baselines} compares answer aggregation on the
same 256 completed traces within each sampling run, at identical token cost.
Self-Consistency
\citep{Wang2022Self} uses majority voting. Self-Certainty
\citep{Kang2025Scalable} uses certainty scores for Best-of-$N$ selection
or rank-weighted voting. Reasoning-Pruning Consistency combines pruning
with probability-weighted voting using its released implementation\footnote{\url{https://github.com/WNJXYK/RPC}}.
Offline DeepConf~\citep{Fu2025Deep} filters completed traces by confidence
before applying confidence-weighted voting to the retained candidate responses.

Self-Certainty achieves $79.63\%$ accuracy with rank-weighted voting,
close to $79.71\%$ for Suffix 500 rank-weighted voting, but $73.85\%$
with Best-of-$N$. The contrast illustrates the importance of
how certainty scores are used. Suffix 500 rank-weighted voting has the highest overall
point estimate among the listed methods and leads on AIME 2025 and
GPQA-Diamond; DeepConf's tail-score variant, which retains $10\%$ of the traces,
leads on AIME 2024.

\begin{table}[!htbp]
  \centering
  \footnotesize
  \caption{Answer aggregation using the same 256 responses and token budget within each sampling run.
  Suffix 500 rank-weighted voting has the highest overall point estimate.
  Accuracy is in percent; differences are percentage points from
  Self-Consistency. Dataset columns average five models; Overall averages
  the 15 settings equally. Bold marks column maxima, including displayed ties.
  The rank exponents are 1.2 for Self-Certainty and 1 for Suffix 500 rank-weighted voting.
  DeepConf retains the indicated fraction of the highest-scoring traces.
  For DeepConf, bottom-group mean averages the lowest-scoring 10\% of sliding
  windows, whereas tail mean averages token confidence over the final segment.}
  \label{tab:tts-fixed-budget-baselines}
  \renewcommand{\arraystretch}{1.06}
  \setlength{\tabcolsep}{0pt}
  \begin{tabular}{@{}p{0.40\linewidth}*{5}{>{\centering\arraybackslash}p{0.12\linewidth}}@{}}
    \toprule
    \textbf{Method} & \shortstack{AIME\\2024} & \shortstack{AIME\\2025} & \shortstack{GPQA-\\Diamond} & Overall & \shortstack{Difference\\(percentage\\points)} \\
    \midrule
    Self-Consistency & 83.41 & 81.40 & 71.32 & 78.71 & +0.00 \\
    Self-Certainty Best-of-$N$ & 80.94 & 71.53 & 69.07 & 73.85 & -4.86 \\
    Self-Certainty: rank-weighted voting & 85.78 & 81.58 & 71.53 & 79.63 & +0.92 \\
    Reasoning-Pruning Consistency & 84.47 & 80.53 & 70.72 & 78.57 & -0.13 \\
    DeepConf: bottom-group mean; retain 10\% & 83.77 & 78.10 & 70.66 & 77.51 & -1.20 \\
    DeepConf: bottom-group mean; retain 90\% & 84.16 & 81.44 & 71.35 & 78.98 & +0.27 \\
    DeepConf: tail mean; retain 10\% & \textbf{86.60} & 79.58 & 71.68 & 79.29 & +0.58 \\
    DeepConf: tail mean; retain 90\% & 84.47 & 81.38 & 71.48 & 79.11 & +0.40 \\
    \midrule
    Suffix 500 rank-weighted voting & 85.59 & \textbf{81.79} & \textbf{71.76} & \textbf{79.71} & \textbf{+1.01} \\
    \bottomrule
  \end{tabular}
\end{table}

We evaluate the adaptive baselines in Table~\ref{tab:tts-adaptive-baselines}
under matched sampling conditions. Adaptive-Consistency
\citep{Aggarwal2023Let} stops sampling based on a probabilistic criterion
for answer consistency. Early-Stopping Self-Consistency~\citep{Li2024Escape}
stops when answers agree within a recent sampling window.
Online DeepConf~\citep{Fu2025Deep} uses confidence thresholds estimated from
warmup traces to truncate low-confidence generations, then applies confidence-weighted voting
over retained traces. DeepConf uses 2,048-token windows or tails; for online
DeepConf, we calibrate thresholds using 16 complete warmup traces for each question.

\begin{table}[!htbp]
  \centering
  \small
  \caption{Accuracy and token cost under fixed and adaptive sampling.
  TACS reaches 79.54\% accuracy at $0.176\times$ the fixed-256
  Self-Consistency cost. All three measures are averaged equally across
  the 15 model--dataset settings. Early-Stopping Self-Consistency allows
  at most 256 traces on AIME and 255 on GPQA-Diamond.
  Retention parameters are 10\% for DeepConf-low and 90\% for DeepConf-high.
  DeepConf counts attempted traces and tokens from truncated
  generations; TACS completes each observed trace.}
  \label{tab:tts-adaptive-baselines}
  \renewcommand{\arraystretch}{1.06}
  \setlength{\tabcolsep}{0pt}
  \begin{tabular}{@{}p{0.36\linewidth}>{\raggedright\arraybackslash}p{0.25\linewidth}*{3}{>{\centering\arraybackslash}p{0.13\linewidth}}@{}}
    \toprule
    \textbf{Method} & \textbf{Sampling budget} & \shortstack{Accuracy\\(\%)} & \shortstack{Relative\\token cost} & \shortstack{Mean trace\\count} \\
    \midrule
    Self-Consistency & 32 traces (fixed) & 78.27 & 0.125 & 32.0 \\
    Self-Consistency & 128 traces (fixed) & 78.63 & 0.500 & 128.0 \\
    Self-Consistency & 256 traces (fixed) & 78.71 & 1.000 & 256.0 \\
    \midrule
    Adaptive-Consistency & At most 256 traces & 78.65 & 0.363 & 26.1 \\
    Early-Stopping Self-Consistency & Dataset-specific maximum & 78.66 & 0.419 & 69.2 \\
    DeepConf-low & At most 256 traces & 76.89 & 0.176 & 57.8 \\
    DeepConf-high & At most 256 traces & 79.00 & 0.555 & 113.6 \\
    \midrule
    \textbf{TACS} & 32 qualifying traces;\newline at most 256 traces & 79.54 & 0.176 & 43.5 \\
    \bottomrule
  \end{tabular}
\end{table}

TACS reaches $79.54\%$ accuracy at relative token cost $0.176$, compared
with $78.65\%$ at $0.363$ for Adaptive-Consistency and $78.66\%$ at $0.419$
for Early-Stopping Self-Consistency. Both DeepConf variants have lower
accuracy at equal or higher displayed token cost.

\FloatBarrier

\section{Reproducibility Details}
\label{app:reproducibility-details}
\subsection{Prompt Templates}

Prompts use a single user message and the model's chat template with a
generation prompt. GPT-OSS-20B uses medium reasoning effort; Gemma models
use thinking mode. Braces indicate placeholders.

\begin{promptbox}{AIME 2024 and AIME 2025}
{\small\ttfamily\raggedright
\{question\}\par
\par
Please reason step by step, and put your final answer within \textbackslash boxed\{\}.}
\end{promptbox}

\begin{promptbox}{GPQA-Diamond}
{\small\ttfamily\raggedright
Answer the following multiple choice question. Please reason step by step, and put your final answer (one of A, B, C, D) within \textbackslash boxed\{\}.\par
\par
\{question\}\par
\par
A) \{choice\_A\}\par
B) \{choice\_B\}\par
C) \{choice\_C\}\par
D) \{choice\_D\}}
\end{promptbox}

\Needspace*{4\baselineskip}
\subsection{Diagnostic Samples and Evaluation Eligibility}
\label{app:diagnostic-samples}

The four-metric comparison and local diagnostics use 512 responses per
model--question pair across five models and three datasets, totaling
660,480 responses. Of these, 641,173 meet the evaluation eligibility
criteria. The token-category and word-position means use all responses;
correctness-based comparisons use the eligible subset. For these analyses,
we extract the final boxed answer and retain responses with a nonempty
answer, as well as any response graded correct.
We refer to retained responses as valid; they include both correct and
incorrect answers. We also repeat the local diagnostics using only valid
responses. Table~\ref{tab:local-group-coverage} in
Appendix~\ref{app:surface-form-diagnostics} reports coverage of the
all-response population used in the main local figures and tables.
In the rank-sensitivity analysis, questions without valid
responses and unestimable question statistics are excluded without imputation.
In the masking
comparisons, all conditions share the same responses, with no
condition-specific exclusions for missing or non-finite scores.

For category-weighting comparisons, both conditions use the same questions
containing Digit tokens for cross-question evaluation and the same responses
containing all four principal categories for within-question evaluation.

For the comparisons in Appendices~\ref{app:target-comparison}--\ref{app:common-question-set},
questions are resampled jointly across models, metrics, and comparisons,
retaining all responses for each sampled question. We recompute all statistics
and equal-weight averages in each replicate, distinguishing same-question
and different-question pairs by original question identity.

\Needspace*{4\baselineskip}
\subsection{Trace Sampling and TACS Simulation}
\label{app:tacs-simulation}

The TACS experiments are offline simulations over pre-sampled trace pools.
For each model--question pair, the main experiments use 1,024 stored responses
and $R=128$ shared replay orders generated with seed 42. Each order contains
256 distinct trace indices.
We apply the stopping and voting rules specified in
Appendix~\ref{app:tts-protocol} to each order: fixed-budget sampling uses
all 256 responses, whereas early allocation observes an initial portion.
Fixed-budget methods share the candidate responses and valid-answer rules
within each replay. Adaptive baselines use the same stored pools and
replay-order construction. TACS voting uses responses with valid answers
and finite scores. Responses with invalid answers still count toward the
32-trace stopping target if their prefix scores qualify.

For the experiments in Section~\ref{sec:test-time-compute}, token cost sums
the complete generated lengths of all observed responses, including those
with invalid answers. Within each setting, relative token cost divides this
total by the corresponding total for fixed 256-trace sampling on matched
orders; reported averages weight these ratios equally, rather than averaging
per-question token ratios. Reported costs describe completion tokens consumed
by the replayed policies.

\begin{table}[!htbp]
  \centering
  \scriptsize
  \setlength{\tabcolsep}{5pt}
  \caption{Key decoding and inference settings used to generate the traces.}
  \label{tab:inference-settings}
  \begin{tabular}{@{}p{0.28\linewidth}p{0.64\linewidth}@{}}
    \toprule
    Parameter & Setting \\
    \midrule
    Temperature & $0.6$ \\
    Top-$p$ & $0.95$ \\
    Maximum generation length & $32{,}768$ tokens for Qwen models; $131{,}072$ tokens for GPT-OSS and Gemma models \\
    Batch size & $256$ traces per vLLM generation call; vLLM maximum concurrent sequences set to $512$ \\
    Number of samples & $1{,}024$ stored responses per model--question pair for the main TACS simulations \\
    Random seed & Offline resampling uses seed $42$; trace generation does not pass an explicit vLLM sampling seed \\
    \bottomrule
  \end{tabular}
\end{table}

\end{document}